\documentclass[10pt,twocolumn,letterpaper]{article}

\usepackage[pagenumbers]{cvpr} 

\usepackage{graphicx}
\usepackage{amsmath}
\usepackage{amssymb}
\usepackage{booktabs}
\usepackage{verbatim}
\usepackage{wrapfig,lipsum,booktabs}
\usepackage[dvipsnames]{xcolor}
\usepackage{mathtools} 
\usepackage{multirow}
\usepackage{algorithm}
\usepackage{xcolor}         
\usepackage{listings}

\usepackage{graphicx}  
\usepackage{color}
\usepackage{colortbl}
\usepackage{comment}

\newcommand{\figref}[1]{Fig. \ref{#1}}

\makeatletter
\def\hlinewd#1{%
\noalign{\ifnum0=`}\fi\hrule \@height #1 \futurelet
\reserved@a\@xhline}

\usepackage{pifont}
\newcommand{\xmark}{\ding{55}}%
\definecolor{recolor}{rgb}{0,1,0}

\definecolor{srcolor}{rgb}{1,0,0}

\definecolor{shcolor}{rgb}{0,0,1}

\usepackage[pagebackref,breaklinks,colorlinks]{hyperref}

\usepackage[capitalize]{cleveref}
\crefname{section}{Sec.}{Secs.}
\Crefname{section}{Section}{Sections}
\Crefname{table}{Table}{Tables}
\crefname{table}{Tab.}{Tabs.}

\def\cvprPaperID{2480                               } 
\def\confName{CVPR}
\def\confYear{2022}

\begin{document}

\title{Cost Aggregation Is All You Need for Few-Shot Segmentation}

\author{Sunghwan Hong$^*$ \\ 
Korea University 
\and
Seokju Cho\thanks{Equal contribution}\\
Yonsei University
\and
Jisu Nam\\
Korea University
\and
Seungryong Kim\thanks{Corresponding author} \\
Korea University
}

\maketitle
\maketitle

\begin{abstract}
We introduce a novel cost aggregation network, dubbed Volumetric Aggregation with Transformers (VAT), to tackle the few-shot segmentation task by using both convolutions and transformers to efficiently handle high dimensional correlation maps between query and support. In specific, we propose our encoder consisting of volume embedding module to not only transform the correlation maps into more tractable size but also inject some convolutional inductive bias and volumetric transformer module for the cost aggregation. Our encoder has a pyramidal structure to let the coarser level aggregation to guide the finer level and enforce to learn complementary matching scores. We then feed the output into our affinity-aware decoder along with the projected feature maps for guiding the segmentation process. Combining these components, we conduct experiments to demonstrate the effectiveness of the proposed method, and our method sets a new state-of-the-art for all the standard benchmarks in few-shot segmentation task. Furthermore, we find that the proposed method attains state-of-the-art performance even for the standard benchmarks in semantic correspondence task although not specifically designed for this task. We also provide an extensive ablation study to validate our architectural choices. The trained weights and codes are available at: ~\url{https://seokju-cho.github.io/VAT/}.
\end{abstract}
\section{Introduction}
Semantic segmentation is one of the fundamental Computer Vision tasks which aims to label each pixel in an image with a corresponding class. With the advent of deep networks and the availability of large-scale datasets with ground-truth segmentation annotations, substantial progress has been made in this task~\cite{long2015fully,noh2015learning,chen2017deeplab,chen2018encoder,tao2020hierarchical}. However, as such advances can be attributed to abundant pixel-wise segmentation maps made by manual annotation, which is often labor-intensive, \textit{few}-shot segmentation task~\cite{ravi2016optimization,shaban2017one} has been introduced to address this, where only a handful of support samples are provided to make a mask prediction for a query, which mitigates the reliance on the labeled data. 

\begin{figure}[t]
\centering
  \includegraphics[width=1\linewidth]{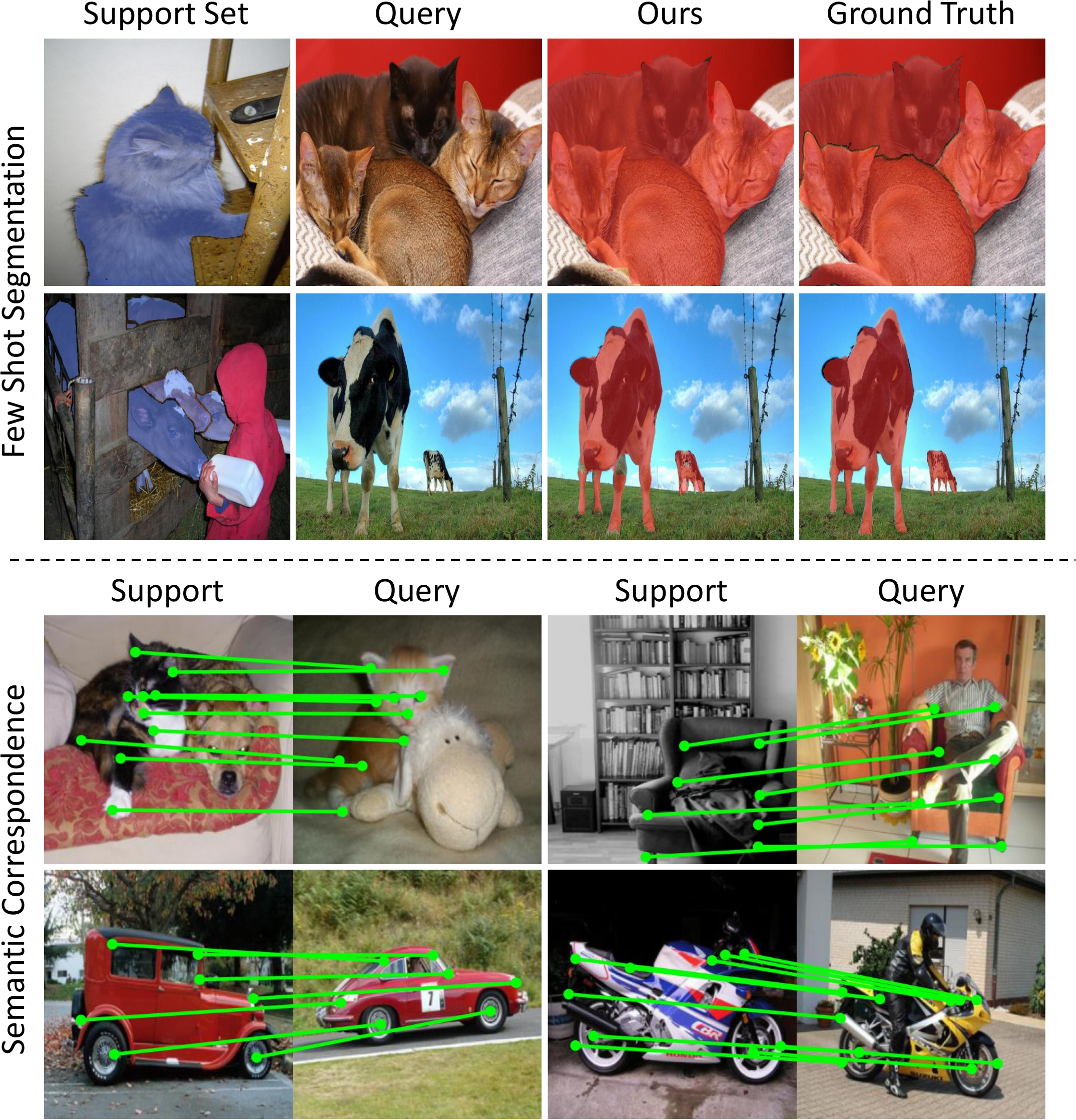}
  \caption{\textbf{Our VAT reformulates few-shot segmentation task to semantic correspondence task.} VAT not only sets state-of-the-art in few-shot segmentation (top), but also attains highly competitive performance for semantic correspondence (bottom).}
\label{fig1}\vspace{-10pt}
\end{figure}

For a decade, numerous methods for few-shot segmentation have been proposed~\cite{wang2020few,yang2020prototype,liu2020part,liu2020crnet,min2021hypercorrelation,lu2021simpler,yang2021mining,xie2021few,wu2021learning,xie2021scale,li2021adaptive,zhang2021self,boudiaf2021few}, and most methods follow a learning-to-learn paradigm~\cite{ravi2016optimization} to avoid the risk of overfitting due to insufficient training data. Since the prediction for query image should be conditioned on support images and corresponding masks, the key to this task is how to effectively utilize provided support samples. Although formulated in various ways, most early efforts~\cite{snell2017prototypical,dong2018few,liu2020part,yang2020prototype} attempted to utilize a prototype extracted from support samples. However, such approaches disregard pixel-wise relationships between support and query features or spatial structure of features, which may lead to sub-optimal results.

In light of this, we argue that few-shot segmentation task can be reformulated as semantic correspondence that aims to find pixel-level correspondences across semantically similar images, which poses some challenges from large intra-class appearance and geometric variations~\cite{ham2016proposal,ham2017proposal,min2019spair}. Recent approaches for this task~\cite{rocco2017convolutional,kim2017fcss,rocco2018end,Rocco18b,min2019hyperpixel,min2020learning,liu2020semantic,truong2020glu,min2021convolutional} carefully designed their models analogously to the classical matching pipeline~\cite{scharstein2002taxonomy,philbin2007object}, i.e., feature extraction, cost aggregation and flow estimation. Especially, the latest works focused on cost aggregation stage~\cite{Rocco18b,min2019hyperpixel,rocco2020efficient,jeon2020guided,liu2020semantic,li2020correspondence,min2021convolutional,cho2021semantic} and showed outstanding performance.

Taking similar approaches, recent few-shot segmentation methods~\cite{zhang2019pyramid,wang2020few,xie2021scale,liu2021few,zhang2021few} also attempted to leverage pixel-wise information by refining features either by using cross-attention~\cite{zhang2021few} or graph attention~\cite{zhang2019pyramid,wang2020few,xie2021scale}. However, as proven in semantic correspondence literature~\cite{min2021convolutional,cho2021semantic}, without aggregating the matching scores, and solely relying on raw correlation maps between features may suffer from the challenges posed due to ambiguities generated by repetitive patterns or background clutters~\cite{rocco2017convolutional,kim2017fcss,lee2019sfnet,truong2020glu,Hong_2021_ICCV}. To address this, one of the latest work, HSNet~\cite{min2021hypercorrelation}, attempts to aggregate the matching scores with 4D convolutions, but it lacks an ability to consider an interaction among the matching scores due to the inherent nature of convolutions. 


In this paper, we introduce a novel cost aggregation network, dubbed Volumetric Aggregation with Transformers (VAT), to tackle the few-shot segmentation task by using both convolutions and transformers to efficiently handle high dimensional correlation maps between query and support. Specifically, within our encoder, we propose volumetric embedding module consisting of a series of 4D convolutions to not only transform high-dimensional correlation maps into more tractable size, but also inject some convolutional bias to aid the subsequent processing by transformers~\cite{xiao2021early}. The output then undergoes volumetric transformer module for the cost aggregation with 4D swin transformer. With these combined, our encoder processes the input in a pyramidal manner to expedite the learning by letting the aggregated correlation maps at coarser level play as a guidance to finer level, enforcing to learn complementary matching scores. Subsequent to encoder, our affinity-aware decoder refines the aggregated costs with the help from appearance affinity and makes a prediction. 

We demonstrate the effectiveness of our method on several benchmarks~\cite{shaban2017one,lin2014microsoft,li2020fss}. Although not specifically designed for semantic correspondence task, our work attains state-of-the-art performance on all the benchmarks for few-shot segmentation and achieves highly competitive results even for semantic correspondence, showing its superiority over the recently proposed methods. We also include a detailed ablation study to justify our choices.

\section{Related Work}
\paragraph{Few-shot Segmentation.} Inspired by few-shot learning paradigm~\cite{ravi2016optimization,snell2017prototypical}, which aims to learn-to-learn a model for a novel task with only a limited number of samples, few-shot segmentation has received considerable attention. Following the success of~\cite{shaban2017one}, prototypical networks~\cite{snell2017prototypical} and numerous other works~\cite{dong2018few,nguyen2019feature,siam2019adaptive,wang2019panet,liu2020part,yang2020prototype,liu2020crnet,xie2021few,yang2021mining,sun2021boosting,zhang2021prototypical,li2021adaptive} proposed to utilize 
a prototype extracted from support samples, which is used to refine the query features to contain the relevant support information. 
In addition, inspired by~\cite{zhang2019canet} that observed the use of high-level features leads to a performance drop,~\cite{tian2020prior} proposed to utilize high-level features by computing a prior map which takes maximum score within a correlation map. Many variants~\cite{sun2021boosting,zhang2021self} extended this idea of utilizing prior maps to guide the feature learning.

\begin{figure*}
    \centering
    \includegraphics[width=1\linewidth]{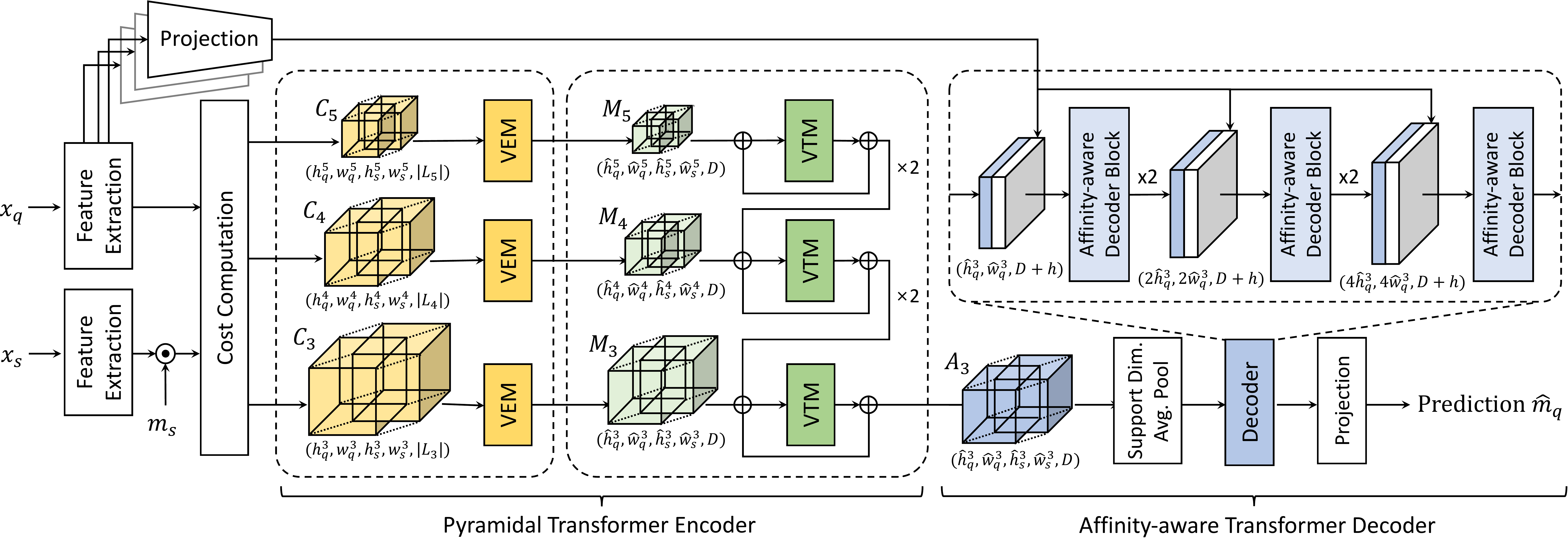}\hfill\\
    \caption{\textbf{Overall network architecture.} Our networks consist of feature extraction and cost computation, pyramidal transformer encoder, and affinity-aware transformer decoder. Given query and support images, we first extract all the intermediate features extracted from backbone network and compute multi-level correlation maps. They then undergo the encoder to aggregate the matching scores with transformers in a pyramidal fashion. The decoder finally predicts a mask label for a query image.}
    \label{fig:overall}\vspace{-10pt}
\end{figure*}

However, as methods based on prototypes or prior maps have apparent limitations, e.g., disregarding pixel-wise relationships between support and query features por spatial structure of feature maps, numerous recent works~\cite{zhang2019pyramid,wang2020few,min2021hypercorrelation,xie2021scale,liu2021few} attempted to fully utilize a correlation map to leverage the pixel-wise relationships between source and query features. Specifically,~\cite{zhang2019pyramid,wang2020few,xie2021scale} use graph attention,~\cite{min2021hypercorrelation} proposes efficient 4D convolutions to fully exploit the multi-level features, and~\cite{liu2021few} formulates the task as optimal transport problem. However, these approaches either do not provide a means to aggregate the matching scores or lack an ability to consider interactions of matching scores. 

As concurrent works,~\cite{zhang2021few} utilizes transformers and proposes to use a cycle-consistent attention mechanism to refine the feature maps to be more discriminative, without considering aggregation of matching scores.~\cite{sun2021boosting} propose global and local enhancement module to refine the features using transformers and convolutions in the decoder, respectively.~\cite{lu2021simpler} focuses solely on the transformer-based classifier by freezing the encoder and decoder, and adapting only the classifier to the task. Unlike them, we take different approach, focusing on aggregation of the high dimensional correlation maps in a novel and efficient way. \vspace{-10pt}  

\paragraph{Semantic Correspondence.}
The objective of semantic correspondence is to find correspondences between semantically similar images with additional challenges posed by large intra-class appearance and geometric variations~\cite{liu2020semantic,cho2021semantic,min2021convolutional}. This is highly similar to few-shot segmentation setting in that few-shot segmentation also aims to label the objects of same class with large intra-class variations, and thus the recent works in both tasks have been taking similar approaches. The latest approaches~\cite{Rocco18b,min2019hyperpixel,rocco2020efficient,jeon2020guided,liu2020semantic,li2020correspondence,min2021convolutional,cho2021semantic} in semantic correspondence focused on cost aggregation stage to find reliable correspondences, and proved its importance. Among those,~\cite{min2021convolutional} proposed to use 4D convolutions for cost aggregation while it showed apparent limitations which include limited receptive fields of convolutions. CATs~\cite{cho2021semantic} resolve this issue and sets a new state-of-the-art by leveraging  transformers~\cite{vaswani2017attention} to aggregate the cost volume. However, it suffers from high computation due to high dimensional nature of correlation map. In this paper, we propose to resolve the aforementioned issues.
\section{Methodology}
\subsection{Problem Formulation}
The goal of \textit{few}-shot segmentation is to segment an object of unseen classes from a query image given only a few annotated examples~\cite{shaban2017one}. To mitigate the overfitting caused by insufficient training data, we follow the common protocol called \textit{episodic} training~\cite{vinyals2016matching}.
Let us denote training and test sets as $\mathcal{D}_\mathrm{train}$ and $\mathcal{D}_\mathrm{test}$, respectively, where object classes of both sets do not overlap. Under $K$-shot setting, multiple \textit{episodes} are formed from both sets, each consisting of a support set $\mathcal{S} = \{(x_{s}^{k}, m_{s}^{k})\}_{k=1}^{\mathrm{K}}$, where $(x_{s}^{k}, m_{s}^{k})$ is $k$-th support image and its corresponding mask pair and a query sample $\mathcal{Q} = (x_q, m_q)$, where $x_q$ and $m_q$ are a query image and mask, respectively. During training, our model takes a sampled episode from $\mathcal{D}_\mathrm{train}$, and learn a mapping from $\mathcal{S}$ and $x_q$ to a prediction $m_q$. At inference, our model predicts $\hat{m}_q$ for randomly sampled $\mathcal{S}$ and $x_q$ from $\mathcal{D}_\mathrm{test}$.

\subsection{Motivation and Overview}
The key to few-shot segmentation is how to effectively utilize provided support samples for a query image. While conventional methods~\cite{tian2020prior,sun2021boosting,zhang2021few,yang2021mining,li2021adaptive} attempted to utilize global- or part-level prototypes extracted from support features, recent methods~\cite{zhang2019pyramid,wang2020few,min2021hypercorrelation,xie2021scale,liu2021few,zhang2021few} attempted to leverage pixel-wise relationships between query and support. One of the latest work, HSNet~\cite{min2021hypercorrelation}, attempts to aggregate the matching scores with 4D convolutions, but it lacks an ability to consider interactions among the matching scores due to the inherent nature of convolutions.    
\begin{figure*}
    \centering
    \includegraphics[width=0.9\linewidth]{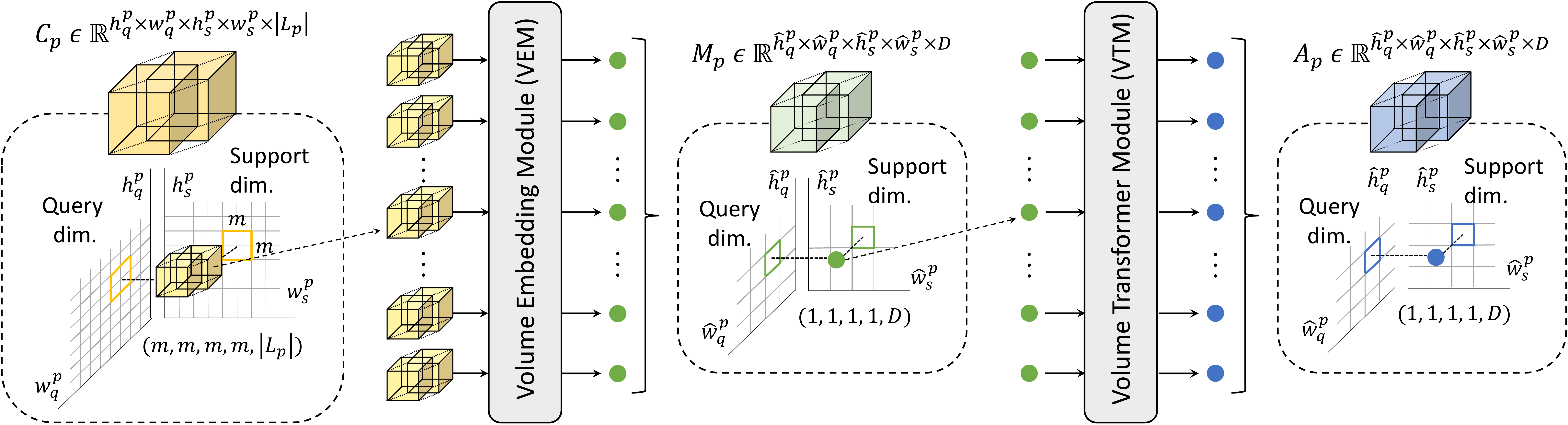}\hfill\\
    \caption{\textbf{Overview of volumetric embedding and transformer modules.} Our VEM aims to ease the subsequent processing of correlation maps by projecting the correlation maps to reduce the spatial dimensions. The output of VEM then undergoes VTM for cost aggregation.}
    \label{fig:vem}\vspace{-10pt}
\end{figure*}

To overcome these, we present a novel design, dubbed volumetric aggregation with transformers (VAT), to effectively integrate information present in all pixel-wise matching costs between query and support with transformers~\cite{vaswani2017attention}. VAT is an encoder-decoder architecture. To compensate for huge complexity of standard transformers~\cite{vaswani2017attention} that prevents from directly applying to correlation maps, for the encoder, we present volume embedding module to effectively reduce the number of tokens while injecting some convolutional inductive bias~\cite{xiao2021early} and volume transformer module based on swin transformer~\cite{liu2021swin}. We design our encoder in a pyramidal fashion to let the output from coarser level cost aggregation to guide the finer level. We then present our decoder that utilizes appearance affinity to resolve the ambiguities in the correlation map and expedite the learning process.

\subsection{Feature Extraction and Cost Computation}\label{sec:3.3}
We first extract features from query and support images and compute initial cost between them. In specific, we follow~\cite{min2021hypercorrelation} to exploit rich semantics present in different feature levels, and build multi-level correlation maps. Given query and support images, $x_q$ and $x_s$, we use CNNs~\cite{he2016deep,simonyan2014very} to produce a sequence of $L$ feature maps, $\{(F_{q}^{l},F_{s}^{l})\}^L_{l=1}$, where $F_{q}^{l}$ and $F_{s}^{l}$ denote query and support feature maps at $l$-th level, respectively. We utilize a support mask, $m_s$, to encode segmentation information and filter out the background information as done in~\cite{li2021adaptive,min2021hypercorrelation,zhang2021self}. We obtain a masked support feature such that $\hat{F}^{l}_{s} = {F}^{l}_{s} \ \odot \ \psi^{l}{(m_{s})}$,
where $\odot$ denotes Hadamard product and $\psi^{l}(\cdot)$ denotes a function that resizes the given tensor followed by expansion along channel dimension of $l$-th layer.

Given a pair of feature maps, $F^l_{q}$ and $F^l_{s}$, we compute a correlation map using the inner product between L-2 normalized features such that
\begin{equation}
    \mathcal{C}^l(i,j)=\mathrm{ReLU}\left(
    \frac{F^l_{q}(i)\cdot \hat{F}^l_{s}(j)}{\|F^l_{q}(i)\|\|\hat{F}^l_{s}(j)\|}\right),
\end{equation}
where $i$ and $j$ denote 2D spatial positions of feature maps, respectively. As done in~\cite{min2021hypercorrelation}, we collect correlation maps computed from all the intermediate features of same spatial size and stack them to obtain a \textit{hypercorrelation} $\mathcal{C}_p = \{\mathcal{C}^l\}_{l\in \mathcal{L}_p} \in \mathbb{R}^{h_{q} \times w_{q} \times h_{s} \times w_{s} \times |\mathcal{L}_p|}$, where $h_{q}, w_{q}$ and $h_{s},  w_{s}$ are height and width of feature maps of query and support, respectively, and $\mathcal{L}_p$ is subset of CNN layer indices $\{1,...,L\}$ at some pyramid layer $p$, indicating the correlation maps of identical spatial size. 

\subsection{Pyramidal Transformer Encoder}
In this section, we show how to effectively aggregate the hypercorrelation with proposed Transformers-based architecture and how to extend this to pyramidal architecture. 
\vspace{-10pt}
\paragraph{Volume Embedding Module.}
Aggregating the hypercorrelation by considering all the query and support spatial dimensions, i.e., $h_{q} \times w_{q} \times h_{s} \times w_{s}$, as \textit{tokens} requires extremely large computation, which has to be resolved to fully exploit all the pixel-wise interactions present in the correlation maps. Perhaps the most straightforward way to reduce the resolutions is to use 4D spatial pooling across query and support spatial dimensions, but this strategy risks losing some information. As an alternative, one can split the hypercorrelation into non-overlapping tensors and embed with a large learnable kernel similarly to a patch embedding in ViT~\cite{dosovitskiy2020image}, but this demands a substantial amount of resources due to the curse of dimensionality. Furthermore, unlike image classification task~\cite{dosovitskiy2020image} which finds a global representation of an image, segmentation aiming for dense prediction needs to consider overlapping neighborhood information.  

To alleviate these limitations, as illustrated in~\figref{fig:vem}, we introduce Volume Embedding Module (VEM) to not only reduce the computation by effectively decreasing the number of tokens, but also inject some convolutional inductive bias~\cite{xiao2021early} to help the subsequent transformer-based model to improve the ability of learning interactions among the hypercorrelation. 
Concretely, we sequentially reduce support and query spatial dimensions by applying 4D spatial max-pooling, overlapping 4D convolutions, RELU, and Group Normalization (GN), where we project the multi-level similarity vector at each 4D position, i.e., projecting a vector size of $|\mathcal{L}_p|$ to a arbitrary fixed dimension denoted as $D$. Considering receptive fields of VEM as 4D window size, i.e., $m\times m\times m\times m$, we build a tensor $\mathcal{M} \in \mathbb{R}^{\hat{h}_{q}\times \hat{w}_{q} \times \hat{h}_{s} \times \hat{w}_{s} \times D}$, where $\hat{h}$ and $\hat{w}$ are the processed sizes. Note that different size of output can be made for source and target spatial dimensions by varying the hyperparameters.

Overall, we define such a process as following:
\begin{equation}
    \mathcal{M}_p = \mathrm{VEM}(\mathcal{C}_p).
    \vspace{-5pt}
\end{equation}

\paragraph{Volumetric Transformer Module.}
Although VEM reduces the number of tokens to some extent, directly applying standard transformers~\cite{dosovitskiy2020image} that has quadratic complexity with respect to number of tokens is still challenging. Our Volumetric Transformer Module (VTM) tackles this by extending swin transformer~\cite{liu2021swin}. By setting a local 4D window, the computation for computing self-attention can be significantly reduced, without significant drop in performance.  While~\cite{wang2020linformer,wu2021fastformer,lu2021soft} could also handle a long sequence of tokens with linear complexity, we argue that as proven in optical flow and semantic correspondence literature~\cite{sun2018pwc,rocco2020efficient} that neighboring pixels tend to have similar correspondences, computing self-attention over local regions and allowing cross-window interaction can help to find reliable correspondences, which in turn yields better segmentation performance in our framework. It should be noted that use of 4D convolutions~\cite{min2021hypercorrelation} also considers local consensus, but it lacks an ability to consider pixel-wise interactions due to the use of fixed kernels during convolution while transformers attentively explore pixel-wise interactions, which we validate our choice in later section.

\begin{figure}
    \centering
    \includegraphics[width=1\linewidth]{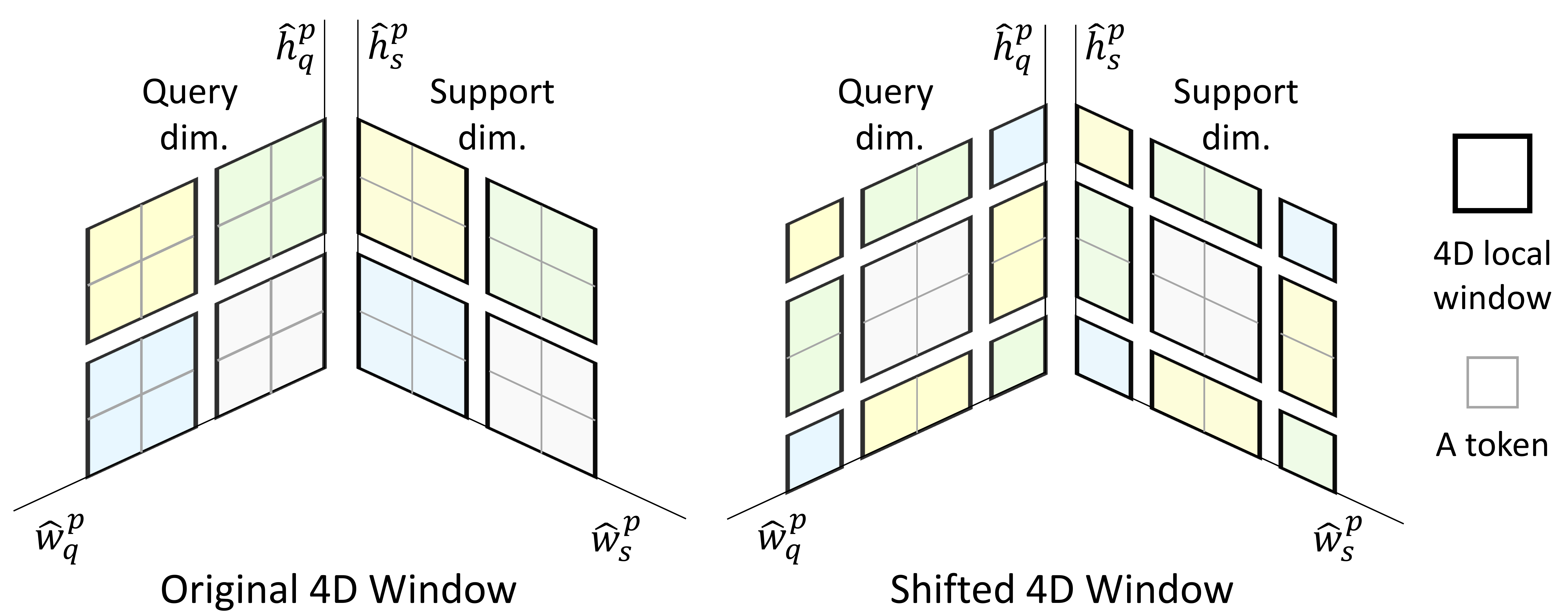}\hfill\\
    \caption{\textbf{Illustration of the shifted 4D window} Our VTM computes self-attention within the partitioned windows in high dimensional space, and it also provides a means to consider inter-window interactions by shifting the 4D window.}    \label{fig:VTM}\vspace{-10pt}
\end{figure}

In specific, for the design of VTM, we extend the original 2D version of swin transformer~\cite{liu2021swin}. 
As illustrated in~\figref{fig:VTM}, we first evenly partition query and support spatial dimensions of $\mathcal{M}_p$ into non-overlapping sub-hypercorrelations $\mathcal{M}'_p \in \mathbb{R}^{n\times n \times n \times n \times D}$. We compute self-attention within each partitioned sub-hypercorrelation. Subsequently, we shift the windows by displacement of $(\lfloor \frac{n}{2} \rfloor,\lfloor \frac{n}{2} \rfloor,\lfloor \frac{n}{2} \rfloor,\lfloor \frac{n}{2} \rfloor)$ pixels from the previously partitioned windows, which we perform self-attention within the newly created windows. Then as done in original swin transformer~\cite{liu2021swin}, we simply roll it back to its original form without adopting any complex implementation. In computing self-attention, we use relative position bias and take the values from an expanded parameterized bias matrix, following~\cite{hu2018relation,hu2019local,liu2021swin}. We leave other components in swin transformer blocks unchanged, e.g., Layer Normalization (LN)~\cite{ba2016layer} and MLP layers.

In addition, to stabilize the learning, we enforce our networks to estimate the residual matching scores as complementary details. We add residual connection in order to expedite the learning process~\cite{he2016deep,cho2021semantic,zhao2021multi}, accounting for fact that at the initial phase when the input $\mathcal{M}_p$ is fed, erroneous matching scores are inferred due to randomly-initialized parameters of transformers, which could complicate the learning process as the networks need to learn the complete matching details from random matching scores.   

To summarize, the overall process is defined as:
\begin{equation}
    \mathcal{A}_p = \mathrm{VTM}(\mathcal{M}_p) = \mathcal{T}(\mathcal{M}_p) + \mathcal{M}_p,
\end{equation}
where $\mathcal{T}$ denotes transformer module.
\vspace{-10pt}

\paragraph{Pyramidal Processing.}
Analogous to~\cite{min2021hypercorrelation,sun2021boosting}, we also utilize the coarse-to-fine approach through a pyramidal processing as illustrated in~\figref{fig:overall}. Note that although it was claimed that utilizing the high-level features has a negative effect on performance~\cite{zhang2019canet}, e.g., conv5$\_$x, numerous recent works~\cite{zhang2021few,min2021convolutional,cho2021semantic,min2021hypercorrelation} in both semantic matching and few-shot segmentation task demonstrated that leveraging multi-level features results in performance boost by large margin. Motivated by this, we also use pyramidal hypercorrelation.

For the coarse-to-fine approach,    
we let the finer level aggregated correlation map $\mathcal{A}_p$ to be guided by the aggregated correlation map of previous (or deeper) levels $\mathcal{A}_{p+1}$. Concretely, aggregated correlation map $\mathcal{A}_{p+1}$ is up-sampled, which is denoted as $\mathrm{up}(\mathcal{A}_{p+1})$, and added to next level's correlation map $\mathcal{A}_p$ to play as a guidance. This process is repeated until the finest level prior to decoder. This differs from~\cite{zhao2021multi,min2021hypercorrelation}, which independently aggregates each correlation map and fuses later, that at each level, they have to learn the matching details from scratch while we are guided by the previous level scores, which dramatically boosts the performance. The pyramidal process is defined as:
\begin{equation}
    \mathcal{A}_p = \mathrm{VTM}(\mathcal{M}_p+\mathrm{up}(\mathcal{A}_{p+1})),
\end{equation}
where $\mathrm{up}(\cdot)$ denotes a bilinear upsampling. 

\subsection{Affinity-Aware Transformer Decoder}
Given the hypercorrelation processed by pyramidal aggregation, we propose to additionally utilize the appearance embedding obtained from query feature maps to effectively decode the hypercorrelation to a query mask. 
From the perspective of correlation maps, this setting can help finding more accurate correspondence as the appearance affinity information helps to filter out the erroneous matching scores, as proven in stereo matching literature, e.g., Cost Volume Filtering (CVF)~\cite{hosni2012fast,sun2018pwc}. Note that HSNet~\cite{min2021hypercorrelation} only decodes the matching costs with a series of 2D convolutions, thus often suffering from ambiguities in the matching costs. 


For the design of our decoder, we first take the average over support dimensions of $\mathcal{A}_p$, which is then concatenated with appearance embedding from query feature maps and processed by swin transformer~\cite{liu2021swin} followed by bilinear interpolation. 
The process is defined as following:
\begin{equation}\label{eq:5}
\hat{\mathcal{A'}} = \mathrm{Decoder}([ \hat{\mathcal{A}},\mathcal{P}(F_{q})]),
\end{equation}
where $\hat{\mathcal{A}} \in \mathbb{R}^{\hat{h}_{q}\times \hat{w}_{q} \times D}$ extracted by average-pooling $\mathcal{A}_p$ in its support spatial dimensions, $\mathcal{P}(\cdot)$ is linear projection, $\mathcal{P}(F_{q})\in \mathbb{R}^{\hat{h}_{q}\times \hat{w}_{q} \times h}$, and $[\,\cdot,\cdot\,]$ denotes concatenation. We sequentially refine the output when immediately after bilinear upsampling to maximize preserving fine details and integrating appearance information. $\hat{\mathcal{A}}$ is bilinearly upsampled and undergoes Eq~\ref{eq:5}  until the projection head which outputs the predicted mask $\hat{m}_q$.

\subsection{Extension to $K$-Shot Setting}
For $K >$ 1, given $K$ pairs of support image and mask $\{(x_{s}^{i}, m_{s}^{i})\}_{i=1}^{\mathrm{K}}$ and a query image $x_q$, our model forward-passes $K$ times to obtain $K$ different query mask $\hat{m}^k_{q}$. We sum up all the $K$ predictions, and find the maximum number of predictions labelled as foreground across all the spatial locations. If the output divided by $k$ is above threshold $\tau$, we label it as foreground, otherwise background.

\begin{table*}
    \begin{center}
    \scalebox{0.8}{
    \begin{tabular}{cl|cccccc|cccccc|c}
            \toprule
            \multirow{2}{*}{\shortstack{Backbone\\feature}} & \multirow{2}{*}{Methods} & \multicolumn{6}{c|}{1-shot} & \multicolumn{6}{c|}{5-shot} & \# learnable \\ 
            
            & & $5^{0}$ & $5^{1}$ & $5^{2}$ & $5^{3}$ &mIoU & FB-IoU & $5^{0}$ & $5^{1}$ & $5^{2}$ & $5^{3}$ &mIoU & FB-IoU & params \\
            \midrule
            

            
            
            \multirow{8}{*}{ResNet50~\cite{he2016deep}} & PANet~\cite{wang2019panet}       & 44.0 & 57.5 & 50.8 & 44.0  &49.1 & - & 55.3 & 67.2 & 61.3 & 53.2  &59.3 & - & 23.5M \\

            & PFENet~\cite{tian2020prior}  & 61.7 & 69.5 & 55.4 & 56.3 &60.8 & 73.3 & 63.1 & 70.7 & 55.8 & 57.9 &61.9 & 73.9 & 10.8M \\ 
            
            & ASGNet~\cite{li2021adaptive} &58.8 &67.9&56.8  &53.7  &59.3 &69.2 &63.4 &70.6 &64.2 &57.4&63.9 &74.2 &10.4M \\
            
            & CWT~\cite{lu2021simpler} &56.3 &62.0 &59.9  &47.2  &56.4 &- &61.3 &68.5 &68.5 &56.6 &63.7  &- &- \\
            & RePRI~\cite{boudiaf2021few}  & 59.8 & 68.3 & \underline{62.1} & 48.5  &59.7 & - & 64.6 & 71.4 & \textbf{71.1} & 59.3 &66.6 & - & - \\ 
            
            & HSNet~\cite{min2021hypercorrelation}   & 64.3 & 70.7 & 60.3 & \textbf{60.5} &64.0 & \underline{76.7} & 70.3 & \underline{73.2} & 67.4 & \textbf{67.1} &\underline{69.5} & \underline{80.6} & \textbf{2.6M} \\
            & CyCTR~\cite{zhang2021few} &\textbf{67.8} &\textbf{72.8} &58.0  &58.0 &\underline{64.2} &-&\underline{71.1} &\underline{73.2} &60.5 &57.5  &65.6  &-&- \\\cline{2-15} \\[-2.0ex]
            & VAT (ours) & \underline{67.6} & \underline{71.2} &\textbf{62.3}  & \underline{60.1}& \textbf{65.3}& \textbf{77.4} & \textbf{72.4} &\textbf{73.6} &\underline{68.6} &\underline{65.7} & \textbf{70.0} &\textbf{80.9} &\underline{3.2M} \\

            \midrule
            
            \multirow{9}{*}{ResNet101~\cite{he2016deep}} & FWB~\cite{nguyen2019feature}        & 51.3 & 64.5 & 56.7 & 52.2 &56.2 & - & 54.8 & 67.4 & 62.2 & 55.3 &59.9 & - & 43.0M \\

            & DAN~\cite{wang2020few}        & 54.7 & 68.6 & 57.8 & 51.6 &58.2 & 71.9 & 57.9 & 69.0 & 60.1 & 54.9 &60.5 & 72.3 & - \\ 
            
            & PFENet~\cite{tian2020prior}  & 60.5 & 69.4 & 54.4 & 55.9 &60.1 & 72.9 & 62.8 & 70.4 & 54.9 & 57.6 &61.4 & 73.5 & 10.8M \\
            & ASGNet~\cite{li2021adaptive}  &59.8 &67.4 &55.6  &54.4 &59.3 &71.7 &64.6 &71.3 &64.2 &57.3   &64.4 &75.2 &10.4M \\
            & CWT~\cite{lu2021simpler}  &56.9 &65.2 &61.2  &48.8 &58.0 &- &62.6 &70.2 &\textbf{68.8}  &57.2&64.7  &- & \\
            & RePRI~\cite{boudiaf2021few}  & 59.6 & 68.6 & \underline{62.2} & 47.2 & 59.4 & - & 66.2 & 71.4 & 67.0 & 57.7 & 65.6 & - & - \\ 
            
            & HSNet~\cite{min2021hypercorrelation}    & 67.3 & 72.3 & 62.0 & \underline{63.1} &\underline{66.2} & \underline{77.6} & 71.8 & \underline{74.4} & 67.0 & \underline{68.3} & \underline{70.4} & \underline{80.6} & \textbf{2.6M} 
            \\
            & CyCTR~\cite{zhang2021few} &\textbf{69.3} &\textbf{72.7} &56.5  &58.6 &64.3 &72.9 &\textbf{73.5} &74.0 &58.6 &60.2 &66.6 &75.0 &- \\\cline{2-15} \\[-2.0ex]
            & VAT (ours) & \underline{68.4}&\underline{72.5} &\textbf{64.8}  & \textbf{64.2} & \textbf{67.5} & \textbf{78.8} & \underline{73.3} &\textbf{75.2} &\underline{68.4} & \textbf{69.5} & \textbf{71.6} & \textbf{82.0}&\underline{3.3M} \\

            \bottomrule
    \end{tabular}
    }
    \vspace{-2.0mm}
    \caption{\textbf{Performance on PASCAL-5$^{i}$~\cite{shaban2017one} in mIoU and FB-IoU.} Numbers in bold indicate the best performance and underlined ones are the second best.}\label{tab:pascal_sota}
    \vspace{-5.0mm}
    \end{center}
\end{table*}

\begin{table*}
    \parbox{.66\linewidth}{
    \centering
        \scalebox{0.63}{
        \begin{tabular}{clcccccccccccc}
                \toprule
                \multirow{2}{*}{\shortstack{Backbone\\feature}} & \multirow{2}{*}{Methods} & \multicolumn{6}{c}{1-shot} & \multicolumn{6}{c}{5-shot} \\ 
                
                & & $20^{0}$ & $20^{1}$ & $20^{2}$ & $20^{3}$ & mean & FB-IoU & $20^{0}$ & $20^{1}$ & $20^{2}$ & $20^{3}$ & mean & FB-IoU \\

                \midrule
                
                \multirow{8}{*}{ResNet50~\cite{he2016deep}} & PMM~\cite{yang2020prototype}        & 29.3 & 34.8 & 27.1 & 27.3 & 29.6 & - & 33.0 & 40.6 & 30.3 & 33.3 & 34.3 & -  \\ 
                
                & RPMM~\cite{yang2020prototype}        & 29.5 & 36.8 & 28.9 & 27.0 & 30.6 & - & 33.8 & 42.0 & 33.0 & 33.3 & 35.5 & -   \\    
                
                & PFENet~\cite{tian2020prior}        & 36.5 & 38.6 & 34.5 & 33.8 & 35.8 & - & 36.5 & 43.3 & 37.8 & 38.4 & 39.0 & - \\   
                & ASGNet~\cite{li2021adaptive} &- &- &-  &- &34.6 &60.4 &- &- &- &- &42.5  &67.0 \\
                & RePRI~\cite{boudiaf2021few}        & 32.0 & 38.7 &  32.7 & {33.1} & 34.1 & - & 39.3 & 45.4 & 39.7 & 41.8 & 41.6 & - \\    
                
                & HSNet~\cite{min2021hypercorrelation}     & 36.3 & \underline{43.1} & 38.7 & 38.7 & 39.2 & \underline{68.2} & \underline{43.3} & \textbf{51.3} & \underline{48.2} & 45.0 & \underline{46.9} & \underline{70.7}\\
                &CyCTR~\cite{zhang2021few} &\underline{38.9} &43.0 &\underline{39.6} &\textbf{39.8} &\underline{40.3} &-  &41.1 &48.9 &45.2 &\textbf{47.0} &45.6 &-\\ \cline{2-14} \\[-2.0ex]
                
                

                
                
                &VAT (ours) &\textbf{39.0} &\textbf{43.8}  &\textbf{42.6} &\underline{39.7} &\textbf{41.3} &\textbf{68.8}  &\textbf{44.1} &\underline{51.1} &\textbf{50.2} &\underline{46.1} &\textbf{47.9} &\textbf{72.4}\\
                
                \bottomrule
        \end{tabular}
        }
        \vspace{-2.0mm}
        \caption{\textbf{Performance on COCO-20$^{i}$~\cite{lin2014microsoft} in mIoU and FB-IoU.}}\label{tab:coco_sota}
        \vspace{-5.0mm}
    }
    \hfill
    \parbox{.30\linewidth}{
    \centering
        
        \scalebox{0.7}{
        \begin{tabular}{clcc}
                \toprule
                \multirow{2}{*}{\shortstack{Backbone\\feature}} & \multirow{2}{*}{Methods} & \multicolumn{2}{c}{mIoU} \\ 
                
                & & 1-shot & 5-shot \\
                
                \midrule

                \multirow{3}{*}{ResNet50~\cite{he2016deep}}       &FSOT~\cite{liu2021few}&82.5&83.8\\
               &HSNet~\cite{min2021hypercorrelation} & \underline{85.5} & \underline{87.8}\\\cline{2-4}\\ [-2.0ex]
                & VAT (ours) &\textbf{89.5} &\textbf{90.3} \\
                \midrule
                \multirow{3}{*}{ResNet101~\cite{he2016deep}} & DAN~\cite{wang2020few} & {85.2} & {88.1} \\ 
                
                & HSNet~\cite{min2021hypercorrelation}  & \underline{86.5} &\underline{88.5} \\ \cline{2-4} \\[-2.0ex]
                & VAT (ours) &\textbf{90.0} & \textbf{90.6} \\

                \bottomrule
        \end{tabular}
        }
        \vspace{-2.0mm}
        \caption{\textbf{Mean IoU comparison on FSS-1000~\cite{li2020fss}.}}\label{tab:fss_sota}
        \vspace{-5.0mm}
    }
\end{table*}

   %
                
  
\section{Experiments}
\subsection{Implementation Details}\label{sec:4.1}
For backbone feature extractor, we use ResNet50 and ResNet101~\cite{he2016deep} pre-trained on ImageNet~\cite{deng2009imagenet}, which are frozen during training, following~\cite{min2021hypercorrelation,zhang2019canet}. We set the threshold $\tau$ to $0.5$. We use data augmentation used in~\cite{cho2021semantic,buslaev2020albumentations} for training. We use AdamW~\cite{loshchilov2017decoupled} with learning rate set to $5\mathrm{e}-4$. We set $D$ to $128$. We use feature maps from conv3$\_$x ($p=3$), conv4$\_$x ($p=4$) and conv5$\_$x ($p=5$) for cost computation. We use last layers from conv2$\_$x, conv3$\_$x and conv4$\_$x for appearance affinity when trained on FSS-1000~\cite{li2020fss} and conv4$\_$x is excluded when trained on PASCAL-5$^{i}$~\cite{shaban2017one} and COCO-20$^{i}$~\cite{lin2014microsoft}. The aforementioned hyperparameters are set with cross-validation. More details can be found in supplementary material. 

\begin{table*}
    \parbox{.31\linewidth}{
    \centering
                \scalebox{0.75}{
        \begin{tabular}{ll|cc}
\toprule
     &\multirow{2}{*}{Components} &\multicolumn{2}{c}{mIoU} \\
 &&1-shot &5-shot \\  \midrule
(\textbf{I})&Baseline&78.7&80.4 \\
(\textbf{II}) &+ VEM& 84.7&85.9 \\
(\textbf{III}) &+ VTM& 86.9& 88.2  \\
(\textbf{IV}) &+ residual connection &\underline{87.0} &\underline{88.4}\\
(\textbf{V}) &+ appearance affinity &\textbf{90.0} &\textbf{90.6} \\\bottomrule
    \end{tabular}
        }
        \vspace{-2.0mm}
        \caption{\textbf{Ablation study of VAT.} }\label{tab:vatable}\vspace{-5.0mm}
    }
    \hfill
    \parbox{.31\linewidth}{
    \centering
        \scalebox{0.75}{
        \begin{tabular}{l|c}
    \toprule
    \multirow{2}{*}{Different aggregators}& FSS-1000~\cite{li2020fss}\\
    &mIoU (\%)\\\midrule
    
       Standard transformer~\cite{vaswani2017attention}&OOM \\
       Center-pivot 4D convolutions~\cite{min2021hypercorrelation}& \underline{88.1}\\
         Linear transformer~\cite{katharopoulos2020transformers}& 87.7 \\
         Fastformer~\cite{wu2021fastformer}& 87.8 \\
         \hline
         Volumetric transformer (ours) &\textbf{90.0} \\
         \bottomrule
    \end{tabular}
        }
        \vspace{-2.0mm}
        \caption{\textbf{Ablation study of cost aggregators.} OOM: Out of Memory.}\label{tab:aggregator}\vspace{-5.0mm}
    }
    \hfill
    \parbox{.31\linewidth}{
    \centering
        
        \scalebox{0.75}{
        \begin{tabular}{l|cc}
\toprule
     \multirow{3}{*}{Backbones feature} &\multicolumn{2}{c}{FSS-1000~\cite{li2020fss}} \\
     &\multicolumn{2}{c}{mIoU ($\%$)}\\
 &1-shot &5-shot \\  \midrule
ResNet50~\cite{he2016deep} &\underline{89.5} &\underline{90.3}\\
ResNet101~\cite{he2016deep} &\textbf{90.0} &\textbf{90.6}\\
PVT~\cite{wang2021pyramid} & \underline{89.5}&89.9\\
Swin transformer~\cite{liu2021swin} &89.1 &89.4 \\\bottomrule
    \end{tabular}
        }
        \vspace{-2.0mm}
        \caption{\textbf{Ablation study of different feature backbone. }}\label{tab:backbone}\vspace{-5.0mm}

    }
\end{table*}
\subsection{Experimental Settings}\label{sec:4.2}
In this section, we conduct comprehensive experiments for few-shot segmentation, by evaluating our approach through comparisons to recent state-of-the-art methods including PMM~\cite{yang2020prototype}, RPMM~\cite{yang2020prototype}, FWB~\cite{nguyen2019feature}, OSLSM~\cite{shaban2017one}, PANet~\cite{wang2019panet}, CANet~\cite{zhang2019canet}, PFENet~\cite{tian2020prior} DAN~\cite{wang2020few}, RePRI~\cite{boudiaf2021few},
SAGNN~\cite{xie2021scale},
FSOT~\cite{liu2021few},
CyCTR~\cite{zhang2021few}, CWT~\cite{lu2021simpler}, ASGNet~\cite{li2021adaptive}, and HSNet~\cite{min2021hypercorrelation}. In Section 4.3, we provide a detailed analysis on our segmentation results evaluated on several benchmarks, and we conduct an extensive ablation study to justify our architectural choices and include an analysis of each component in Section 4.4.
\vspace{-10pt}

\paragraph{Datasets.}
We evaluate our approach on three standard few-shot segmentation datasets, PASCAL-5$^{i}$~\cite{shaban2017one}, COCO-20$^{i}$~\cite{lin2014microsoft}, and FSS-1000~\cite{li2020fss}. PASCAL-5$^{i}$ is created from images from PASCAL VOC 2012~\cite{everingham2010pascal} and extra mask annotations~\cite{hariharan2014simultaneous}, Originally, 20 object classes are available, but for cross-validation, as done in OSLM~\cite{shaban2017one}, they are evenly divided into 4 folds $i \in \{0,1,2,3\}$, and this makes each fold contains 5 classes. COCO-20$^{i}$ contains 80 object classes, and as done for PASCAL-5$^{i}$, the dataset is evenly divided into 4 folds, which results 20 classes for each fold. FSS-1000 is a more diverse dataset consisting of 1000 object classes. Following~\cite{li2020fss}, we divide 1000 categories into 3 splits for training, validation and testing, which consist of 520, 240 and 240 classes, respectively. For PASCAL-5$^{i}$ and COCO-20$^{i}$, we follow the common evaluation   practice~\cite{min2021hypercorrelation,tian2020prior,liu2020part} and standard cross-validation protocol; for each $i$-th fold, the target fold $i$ is used for evaluation and other folds are used for training.
\vspace{-10pt}

\paragraph{Evaluation Metric.} 
Following common practice~\cite{zhang2019canet,tian2020prior,min2021hypercorrelation,zhang2021few}, we adopt mean intersection over union (mIoU) and foreground-background IoU (FB-IoU) as our evaluation metric. The mIoU averages over all IoU values for all object classes such that $\mathrm{mIoU} = \frac{1}{C}\sum _{c=1}^{C} \mathrm{IoU}_{c}$, where $C$ is the number of classes in each fold, e.g., $C = 20$ for COCO-20$^{i}$. FB-IoU disregards the object classes and averages over foreground and background IoU, $\mathrm{IoU}_\mathrm{F}$ and $\mathrm{IoU}_\mathrm{B}$, such that $\mathrm{FBIoU} = \frac{1}{2}(\mathrm{IoU}_\mathrm{F} + \mathrm{IoU}_\mathrm{B})$. As stated in~\cite{zhang2019canet}, we mainly focus on mIoU since it better accounts for generalization power of a model than FB-IoU.

\subsection{Segmentation Results }
\begin{figure}
    \centering
    \includegraphics[width=1\linewidth]{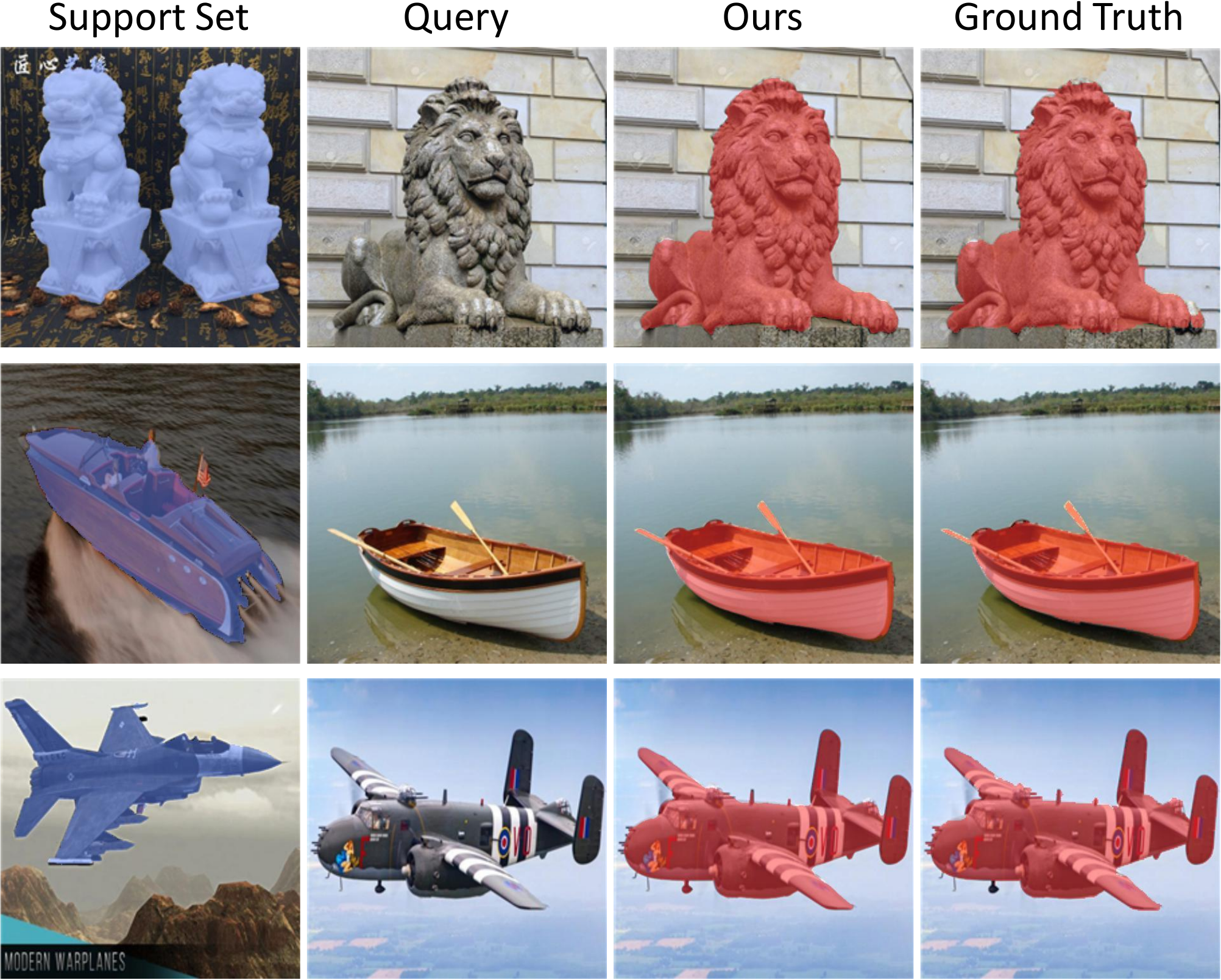}
    \caption{\textbf{Qualitative results on FSS-1000~\cite{li2020fss}.}}
    \label{fig:qualifss}\vspace{-10pt}
\end{figure}

Table~\ref{tab:pascal_sota} summarizes quantitative results on the PASCAL-5$^{i}$~\cite{shaban2017one}. We denote the type of backbone feature extractor and number of parameters for comparison. For PASCAL-5$^{i}$, we tested with two backbone networks, ResNet50 and ResNet101~\cite{he2016deep}. The proposed method outperforms other methods for almost all the folds with respect to both mIoU and FB-IoU. Consistent with this, VAT also attains state-of-the-art performance on COCO-20$^{i}$~\cite{lin2014microsoft} for both 1-shot and 5-shot. Interestingly, for the most recently introduced dataset deliberately created for few-shot segmentation task, FSS-1000~\cite{li2020fss}, VAT outperforms HSNet~\cite{min2021hypercorrelation} and FSOT~\cite{liu2021few} by a large margin, almost 4$\%$ increase in mIoU compared to HSNet with ResNet50. Overall, VAT sets a new state-of-the-art for all the benchmarks and this is confirmed at both Table~\ref{tab:coco_sota} and Table~\ref{tab:fss_sota}. The qualitative results are shown in~\figref{fig:qualifss}.

Another remarkable point is that, we find that even with larger number of learnable parameter than~\cite{min2021hypercorrelation}, our method still outperforms. As shown in Table~\ref{tab:pascal_sota}, the models with larger number of parameters tend to perform worse. It is well known that the number of parameters has inverse relation to generalization power~\cite{Tetko1995NeuralNS}. However, VAT successfully avoids this by focusing on cost aggregation, showing that simply reducing the number of parameters may not be the answer.


\subsection{Ablation study}
In this section, we show an ablation analysis to justify the architectural choices we made, explore impact of feature backbone and investigate whether VAT is transferable to semantic correspondence task. Throughout this section, all the experiments are conducted on FSS-1000~\cite{li2020fss} datasets with ResNet-101 as backbone feature extractor unless specified. Each ablation experiment is conducted under same experimental setting for a fair comparison. For semantic correspondence ablation study, we use percentage of correct keypoints (PCK) as our evaluation metric.
\vspace{-10pt}

\paragraph{Effectiveness of each component in VAT. }
We consider these as core components: volume embedding module, volumetric transformer module, residual connection and appearance affinity. We define the baseline for this ablation study as a model that replaces VEM with simple 4D max-pooling, skips VTM to show the importance of cost aggregation, excludes residual connection around cost aggregation module and disregards appearance affinity. We evaluate this by adding components progressively. 

As summarized in Table~\ref{tab:vatable}, each component helps to boost the performance. Especially, from (\textbf{I}) to (\textbf{III}) and (\textbf{IV}) to (\textbf{V}), we observe significant improvement. We find that leveraging appearance affinity helps to find accurate correspondences, which in turn yields better segmentation performance, and VEM followed by VTM not only eases the computational burden but also boosts the performance. \vspace{-10pt}

\paragraph{Is VTM better than other aggregators?}
As summarized in Table~\ref{tab:aggregator}, we provide an ablation study to justify the use of our pyramidal transformers for cost aggregation. To this end, it is perhaps necessary to compare with other methods including 4D convolutions~\cite{Rocco18b,min2021hypercorrelation} and efficient transformers~\cite{wu2021fastformer,wang2020linformer}. As 4D convolution as in~\cite{min2021hypercorrelation} also aggregates the matching scores in a local manner similar to swin transformer~\cite{liu2021swin}, we conjecture that it should perform similarly or even better. We investigate this and report the results to confirm that VTM exceeds other aggregators. 

For a fair comparison, we only replace VTM with other aggregators and leave all the other components in our architecture unchanged. We observe that our proposed method outperforms other aggregators by a large margin. Note that the use of standard transformers for cost aggregation is inapplicable due to high dimensional space of hypercorrelation. Interestingly, although center-pivot 4D convolutions~\cite{min2021hypercorrelation} also focus on locality as swin transformer~\cite{liu2021swin}, the performance gap indicates that the ability to attentively consider pixel-wise interactions during the self-attention computation is critical. Another interesting point is that linear transformer~\cite{katharopoulos2020transformers} and fastformer~\cite{wu2021fastformer} that exploit from global receptive fields of transformers achieve similar performance. This confirms that for the cost aggregation, our approach may be more suitable. \vspace{-10pt} 

\begin{table}[]
    \centering
    \scalebox{0.75}{
          \begin{tabular}{lccc}
        \hlinewd{0.9pt}
        \multirow{2}{*}{Methods} &SPair-71k~\cite{min2019spair}&PF-PASCAL~\cite{ham2017proposal}&PF-WILLOW~\cite{ham2016proposal} \\
        &$\alpha_{\text{bbox}} = 0.1$&$\alpha_{\text{img}}$  = 0.1 &$\alpha_{\text{bbox}}$ = 0.1\\
        \midrule
        SCOT~\cite{liu2020semantic}  &35.6&85.4&76.0 \\
        CHM~\cite{min2021convolutional} &46.3  &{91.6}&\underline{79.4} \\
        MMNet~\cite{zhao2021multi}  &\underline{50.4}&{91.6}&- \\
        PMNC~\cite{Lee_2021_CVPR}  &\underline{50.4}&90.6&- \\
        CATs~\cite{cho2021semantic} &49.9 &\textbf{92.6} &79.2\\
        \hline
        VAT (ours)  &\textbf{54.2}&\underline{92.3}&\textbf{81.0} \\
        
        \hlinewd{0.8pt}
\end{tabular}}
\vspace{-5pt}
    \caption{\textbf{Quantitative results on SPair-71k~\cite{min2019spair}, PF-PASCAL~\cite{ham2017proposal} and PF-WILLOW~\cite{ham2016proposal}.}}
    \label{tab:semantic}\vspace{-10pt}
\end{table}
\paragraph{Does different feature extractor matter?}
Conventional few-shot segmentation methods only utilized CNN-based feature backbones~\cite{he2016deep} for extracting features.~\cite{zhang2019canet} observed that high-level features contains semantics of objects which could lead to overfitting and not suitable to use for the task of few-shot segmentation. Then the question naturally arises. What about transformer-based backbone networks? As addressed in many works~\cite{raghu2021vision,dosovitskiy2020image}, CNN and transformers see images differently, which means that the kinds of backbone networks may affect the performance significantly, but this has never been explored in this task. We thus exploit several well-known vision transformer architectures to explore the potential differences that probably exist.

The results are summarized in Table~\ref{tab:backbone}. We were surprised to find that both convolution- and transformer-based backbone networks attain similar performance. We conjecture that although it has been widely studied that convolutions and transformers see differently~\cite{raghu2021vision}, as they are pre-trained on the same dataset~\cite{deng2009imagenet}, the representations learned by models are almost alike. Note that we only utilized backbones with pyramidal structure, and the results may differ if other backbone networks are used, which we leave this exploration for the future work.
\vspace{-10pt}


\paragraph{Can VAT also perform well in semantic correspondence task?}
To tackle the few-shot segmentation task, we reformulated it as finding semantic correspondences under conditions of several challenges posed by large intra-class variations and geometric deformations. This means that the proposed method should be able to perform well in semantic correspondence task at least to some extent. Here, we compare our method with other state-of-the-art methods in semantic correspondence task and investigate the key aspect: To what extent and how well can VAT perform in semantic correspondence task given similar formulation?

\begin{figure}[t]
\centering
    \begin{subfigure}{0.495\linewidth}{\includegraphics[width=1\linewidth]{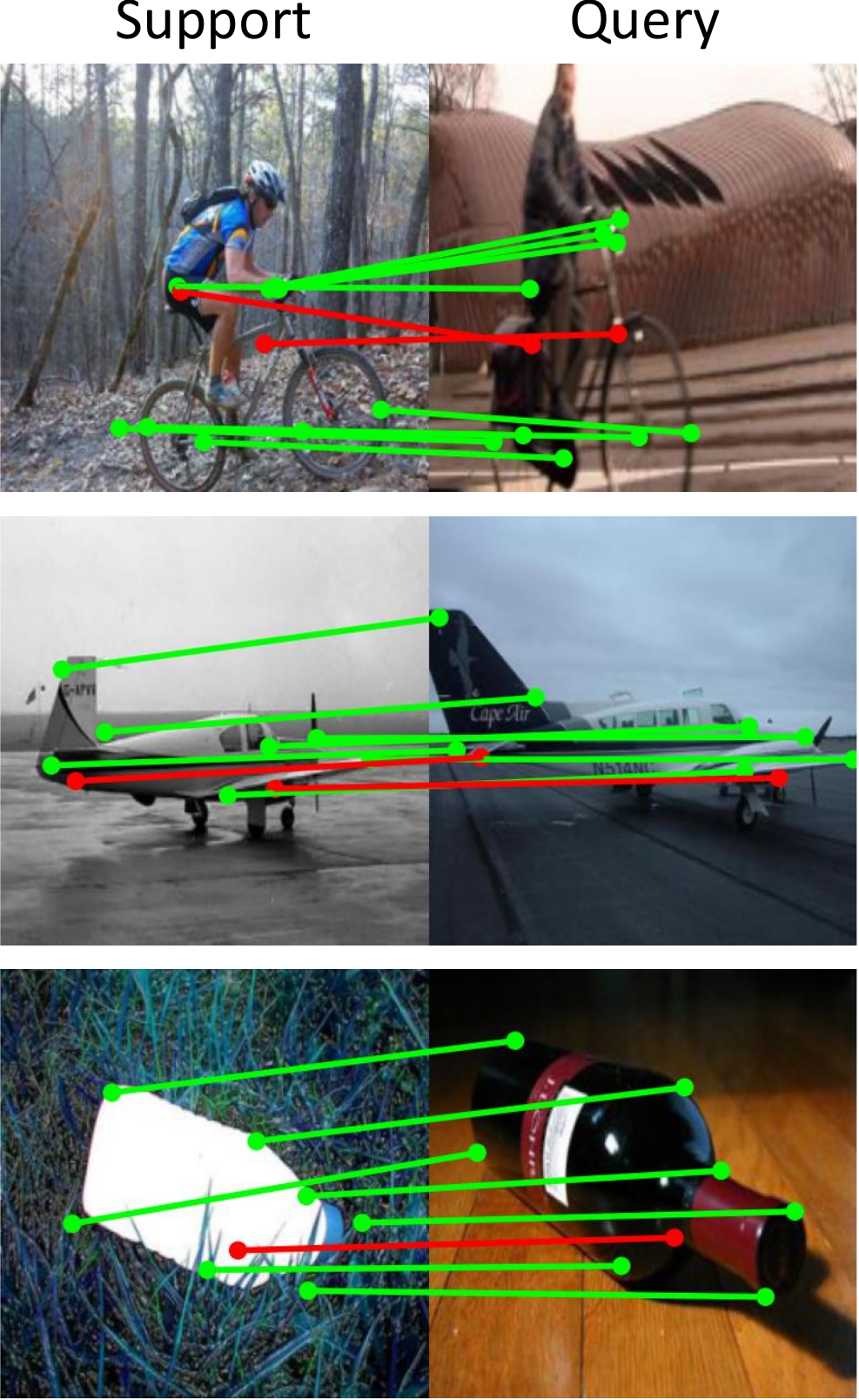}}\caption{CATs~\cite{cho2021semantic}}\end{subfigure}\hfill
    \begin{subfigure}{0.495\linewidth}{\includegraphics[width=1\linewidth]{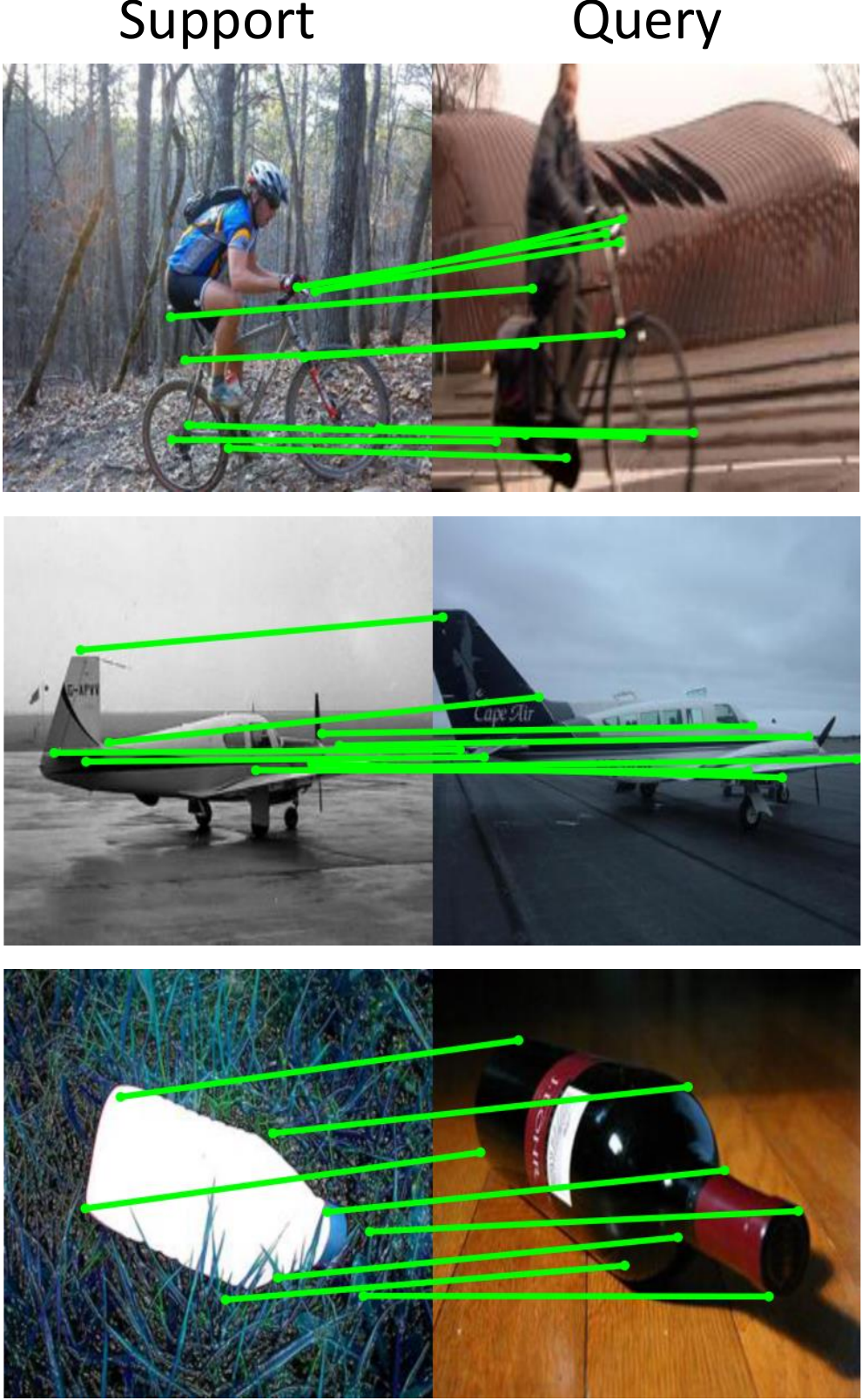}}\caption{Ours}\end{subfigure}\hfill\\
    \vspace{-5pt}
   \caption{\textbf{Qualitative results on PF-PASCAL~\cite{ham2017proposal}.}}
   \label{fig:pfpascal}\vspace{-10pt}
\end{figure}

For this ablation study, we made minor modifications to our model: {\it i.e.,} output spatial resolution and embedded features. We refer the readers to either Appendix or github page:~\url{https://seokju-cho.github.io/VAT/} for details. Note that this is possible as our projection head also outputs the same dimension as if a flow field is inferred. Following common protocol~\cite{min2019hyperpixel,huang2019dynamic,min2021convolutional,cho2021semantic}, we use standard benchmarks for this task, which we trained our model on training split of PF-PASCAL~\cite{ham2017proposal} when evaluated on test split of PF-PASCAL~\cite{ham2017proposal} and PF-WILLOW~\cite{ham2016proposal}, and trained on SPair-71k~\cite{min2019spair} when evaluated on SPair-71k~\cite{min2019spair}.  We fine-tuned the feature backbone for fair comparison. As shown in Table~\ref{tab:semantic} and~\figref{fig:pfpascal}, although not specifically designed for semantic correspondence task, VAT either sets a new state-of-the-art~\cite{min2019spair,ham2016proposal} or attain the second highest PCK~\cite{ham2017proposal} for this task. This results show that cost aggregation is a prime importance in both few-shot segmentation and semantic correspondence.

\section{Conclusion}
In this paper, we have proposed, for the first time, to aggregate the high dimensional correlation maps with both convolutions and transformers for few-shot segmentation. For our pyramidal encoder, to handle the high computation caused by direct aggregation of correlation maps, we introduced volume embedding module, and for the cost aggregation, we proposed volumetric transformer module. Then the output undergoes the proposed affinity-aware decoder for prediction. We have shown that although the proposed method is not specifically designed for semantic correspondence task, it attains state-of-the-art performance for all the standard benchmarks for both few-shot segmentation and semantic correspondence. Also, we have conducted extensive ablation study.

\medskip
{\small
\bibliographystyle{plain}
\bibliography{egbib}
}
\newpage
\clearpage
\begin{center}
	\textbf{\Large Appendix}
\end{center}

\appendix
\renewcommand{\thefigure}{\arabic{figure}}
\renewcommand{\theHfigure}{A\arabic{figure}}
\renewcommand{\thetable}{\arabic{table}}
\renewcommand{\theHtable}{A\arabic{table}}
\setcounter{figure}{0}
\setcounter{table}{0}

In this document, we provide detailed implementation details, more quantitative results on semantic correspondence benchmarks, including SPair-71k~\cite{min2019spair}, PF-PASCAL~\cite{ham2017proposal}, and PF-WILLOW~\cite{ham2016proposal}, and more qualitative results on all the benchmarks we used.

\section*{Appendix A. Implementation details}
\paragraph{Implementation Details.}
We first extract backbone features using ResNet50~\cite{he2016deep} and ResNet101~\cite{he2016deep} from conv3$\_$x, conv4$\_$x and conv5$\_$x. This feature extracting scheme results in 3 pyramidal layers. To compose VEM, we stack a series of $3\times 3 \times 3 \times 3$ convolutions followed by ReLU and Group Normalization. The coarsest level correaltion map has a size of $8\times 8 \times 8 \times 8$ after undergoing VEM. Finer levels, $p=4$ and $p=3$, have volume embeddings of size $16\times 16 \times 8 \times 8$ and $32 \times 32 \times 8 \times 8$, respectively. We set the number of heads in VTM as 4. At $p=5$ level, we set the depth of VTM as 4 and rest are set to 2. Window size $M$ is set to 4 at all levels of VTM. We project the feature maps of query at last layer of level $p=4, p=3$ and $p=2$ to have channels 64, 32 and 16 from 1024, 512 and 256, respectively. We implemented our work with PyTorch~\cite{paszke2017automatic}. We used learning rate of 5e-4. We used augmentation introduced in~\cite{buslaev2020albumentations} for training as shown in Table~\ref{tab:augmentation}. We adopted early-stopping for all the training, which we picked the model with best validation score. We set the batch size as 16.
\vspace{-10pt}

\paragraph{Components of VAT.}
For the ablation study of components of VAT, we defined the baseline model which substitutes VEM with 4D max-pooling. As VTM is a transformer module that takes an input and outputs the same shape to input, it can simply be skipped. Residual connections are simply removed to test its efficiency, and appearance affinity is excluded at the decoder. This experiment is conducted in a sequential manner, where each component is added progressively. 
\vspace{-10pt}

\paragraph{Different Cost Aggregators.}
For the ablation study of cost aggregators, we used codes released by authors of works, including standard transformer~\cite{vaswani2017attention}, center-pivot 4D convolutions~\cite{min2021hypercorrelation}, linear transformer~\cite{katharopoulos2020transformers}, and fastformer~\cite{wu2021fastformer}. We simply replaced VTM with those in order to make a fair comparison.\vspace{-10pt}

\paragraph{Different backbone networks.} Ablation study on different feature backbone follows similar procedure, as PVT~\cite{wang2021pyramid} and swin transformer~\cite{liu2021swin} have pyramidal architecture that outputs features of size and dimensions similar to ResNet~\cite{he2016deep}. 
\vspace{-10pt}

\begin{table*}[]
    \begin{center}
        \scalebox{0.75}{
        \begin{tabular}{l|cccccccccccccccccc|c}
        \toprule
         Methods & aero. & bike & bird & boat & bott. & bus & car & cat & chai. & cow & dog & hors. & mbik. & pers. & plan. & shee. & trai. & tv & all\\
        \midrule

        CNNGeo~\cite{rocco2017convolutional} &  23.4 & 16.7 & 40.2 & 14.3 & 36.4 & 27.7 & 26.0 & 32.7 & 12.7 & 27.4 & 22.8 & 13.7 & 20.9 & 21.0 & 17.5 & 10.2 & 30.8 & 34.1 & 20.6  \\
      
       WeakAlign~\cite{rocco2018end} &  22.2 & 17.6 & 41.9 & {15.1} & {38.1} & {27.4} & {27.2} & 31.8 & 12.8 & 26.8 & 22.6 & 14.2 & 20.0 & 22.2 & 17.9 & 10.4 & {32.2} & 35.1 & 20.9 \\
    
       NC-Net~\cite{Rocco18b} & 17.9 & 12.2 & 32.1 & 11.7 & 29.0 & 19.9 & 16.1 & 39.2 & 9.9 & 23.9 & 18.8 & 15.7 & 17.4 & 15.9 & 14.8 & 9.6 & 24.2 & 31.1 & 20.1   \\\\[-2.3ex]

        HPF~\cite{min2019hyperpixel} & {25.2} & {18.9} & {52.1} & {15.7} & {38.0} & {22.8} & {19.1} & {52.9} & {17.9} & {33.0} & {32.8} & {20.6} & {24.4} & {27.9} & 21.1 & {15.9} & {31.5} & {35.6} & {28.2} \\\\[-2.3ex]
        
        {SCOT}~\cite{liu2020semantic} & {34.9} & {20.7} & {63.8} & {21.1} & {43.5} & {27.3} & {21.3} & {63.1} & {20.0} & {42.9} & {42.5} & {31.1} & {29.8} & {35.0} & {27.7} & {24.4} & {48.4} & {40.8} & {35.6} \\
       
       {DHPF}~\cite{min2020learning} & {38.4} & {23.8} & {68.3} & {18.9} & {42.6} & {27.9} & {20.1} & {61.6} & {22.0} & {46.9} & {46.1} & {33.5} & {27.6} & {40.1} & {27.6} & {28.1} & {49.5} & {46.5} & {37.3} \\

        {CHM}~\cite{min2021convolutional} & {49.6} & {29.3} & {68.7} & {29.7} & {45.3} & {48.4} & {39.5} & {64.9} & {20.3} & {60.5} & {56.1} & {46.0} & {33.8} & {44.3} & {38.9} & {31.4} & {72.2} & {55.5} & {46.3}  \\
        MMNet~\cite{zhao2021multi}  &\underline{55.9}  &\underline{37.0}  &65.0  &35.4 &\underline{50.0}  &\textbf{63.9}  &\textbf{45.7} &62.8  &\textbf{28.7}  &65.0  &54.7  &51.6  &38.5  &34.6  &41.7  &\underline{36.3}  &\textbf{77.7}  &\underline{62.5}  &\underline{50.4}  \\
        PMNC~\cite{Lee_2021_CVPR}  &54.1  &35.9  &\textbf{74.9}  &\underline{36.5} &42.1  &48.8  &40.0 &\textbf{72.6}  &21.1  &\underline{67.6}  &\underline{58.1}  &50.5  &40.1  &\textbf{54.1}  &\underline{43.3}  &35.7  &\underline{74.5}  &59.9  &\underline{50.4}  \\
        CATs~\cite{cho2021semantic}  & {52.0} & {34.7} & {72.2} & {34.3} & {49.9} & {57.5} & \underline{43.6} & {66.5} & \underline{24.4} & {63.2} & {56.5} & \underline{52.0} & \underline{42.6} & {41.7} & {43.0} & {33.6} & {72.6} & {58.0} & {49.9} \\\midrule
        VAT (ours)  &\textbf{56.5}& \textbf{37.8}& \underline{73.0}& \textbf{38.7} &\textbf{50.9} &\underline{58.2}  &40.8  &\underline{70.5} &20.4  &\textbf{72.6}  &\textbf{61.1}  &\textbf{57.8}  &\textbf{45.6}  &\underline{48.1}  &\textbf{52.4}  &\textbf{39.7}  &\textbf{77.7}  &\textbf{71.4}  &\textbf{54.2}  \\
       \bottomrule
        \end{tabular}} 
    \caption{\label{tab:spair} \textbf{Per-class quantitative evaluation on SPair-71k~\cite{min2019spair} benchmark.}}\vspace{-10pt}
    
    \end{center}\vspace{-10pt}
\end{table*}
\begin{table}[]

    \begin{center}
    
    \scalebox{0.8}{
    \begin{tabular}{l|cc|cc}
            \toprule
             \multirow{3}{*}{Methods}& \multicolumn{2}{c|}{PF-PASCAL~\cite{ham2017proposal}} & \multicolumn{2}{c}{PF-WILLOW~\cite{ham2016proposal}}  \\
          
              & \multicolumn{2}{c|}{PCK @ $\alpha_{\text{img}}$}  & \multicolumn{2}{c}{PCK @ $\alpha_{\text{bbox}}$}   \\ 
        
             & 0.1 & 0.15  & 0.1 & 0.15  \\ 
             \midrule
        
             CNNGeo~\cite{rocco2017convolutional}  &69.5 &80.4 &69.2 &77.8\\
             
              WeakAlign~\cite{rocco2018end}  &74.8 &84.0 &70.2 &79.9\\
              
              SFNet~\cite{lee2019sfnet} &81.9 &90.6  &74.0 &84.2\\

             NC-Net~\cite{Rocco18b} &78.9 &86.0  &67.0 &83.7\\
             
             DCC-Net~\cite{huang2019dynamic}  &82.3 &90.5  &73.8 &86.5\\
             
             HPF~\cite{min2019hyperpixel}  &84.8 &92.7 &74.4 &85.6\\
             GSF~\cite{jeon2020guided} &87.8 &95.9  &78.7 &90.2\\
           
             DHPF~\cite{min2020learning}   &90.7 &95.0  &77.6 &89.1 \\
             SCOT~\cite{liu2020semantic}   &85.4 &92.7 &76.0 &87.1\\
             CHM~\cite{min2021convolutional}   &{91.6} &94.9  &\underline{79.4} &87.5\\
             MMNet~\cite{zhao2021multi}  &{91.6} &95.9 &-&-\\
             PMNC~\cite{Lee_2021_CVPR}  &90.6 &- &-&-\\
             CATs~\cite{cho2021semantic}  &\textbf{92.6} &\textbf{96.4} &{79.2} &\underline{90.3}\\\midrule
             VAT (ours) &\underline{92.3} &\underline{96.1} &\textbf{81.0}&\textbf{91.4}\\\bottomrule

    \end{tabular}
    }\caption{\textbf{Quantitative evaluation on standard benchmarks~\cite{min2019spair,ham2016proposal,ham2017proposal}.} 
    }\label{tab:pascal}\vspace{-5pt}
    \end{center}
\end{table}
\begin{table}[]
    \centering
        \scalebox{0.75}{
        \begin{tabular}{l|cc}
    \toprule
    \multirow{2}{*}{Different aggregators}& Memory &Run-time\\
    &(GB) &(ms)\\\midrule
    
       Standard transformer~\cite{vaswani2017attention}&OOM&N/A \\
       Center-pivot 4D convolutions~\cite{min2021hypercorrelation}& \textbf{3.5}
       &\textbf{52.7}\\
         Linear transformer~\cite{katharopoulos2020transformers}& \textbf{3.5}&\underline{56.8} \\
         Fastformer~\cite{wu2021fastformer}&\textbf{3.5} &122.9 \\
         \hline
         Volumetric transformer (ours) &\underline{3.8}&57.3 \\
         \bottomrule
    \end{tabular}
        }
    \caption{\textbf{Memory and Run-time comparison between different aggregators.} The memory and run-time are measured by simply replacing the aggregator module within VAT.}
    \label{tab:memory}
\end{table}

\begin{table}[]
    \centering
        \scalebox{0.75}{
        \begin{tabular}{l|cc}
    \toprule
    \multirow{2}{*}{}& Memory &Run-time\\
    &(GB) &(ms)\\\midrule
    
       HSNet~\cite{min2021hypercorrelation}&\textbf{2.2}&\textbf{51.7} \\
         \hline
         VAT (ours) &\underline{3.8}&\underline{57.3}\\
         \bottomrule
    \end{tabular}
        }
    \caption{\textbf{Memory and Run-time comparison to HSNet~\cite{min2021hypercorrelation}.}}
    \label{tab:overallmemory}
\end{table}
\paragraph{VAT in Semantic Correspondence.}

For the experiments to validate effectiveness of VAT on semantic correspondence, we used the three most widely used datasets in this task, which include SPair-71k~\cite{min2019spair}, PF-PASCAL~\cite{ham2017proposal} and PF-WILLOW~\cite{ham2016proposal}. We made minor modifications to our model: we excluded bilinear upsampling in the affinity-aware decoder, which the original output shape is $128 \times 128$ but with the modification, now the output shape is $32 \times 32$ as two x2 bilinear upsampling is excluded. We used embedded features from conv4$\_$ and conv5$\_$x, which is different from the training setting for the experiments on few-shot segmentation datasets. Note that we do not utilize segmentation mask for this task. We only use keypoints provided in the datasets for training similarly to~\cite{cho2021semantic}. We used the same data augmentation scheme~\cite{cho2021semantic, buslaev2020albumentations} as Table~\ref{tab:augmentation}. We used learning rate of 3e-6 for the feature backbone and 3e-4 for VAT. We used batch size of 8. For the evaluation, we followed the standard protocol~\cite{min2019hyperpixel,min2020learning,liu2020semantic,jeon2020guided,cho2021semantic,min2021convolutional}.
\begin{table}[]
    \centering
    \scalebox{0.75}{
    \begin{tabular}{cl|c}
       \toprule
        &Augmentation type &Probability \\
        \midrule
        \textbf{(I)} &ToGray  &0.2 \\
        \textbf{(II)} &Posterize &0.2 \\
        \textbf{(III)} &Equalize &0.2 \\
        \textbf{(IV)} &Sharpen &0.2 \\
        \textbf{(V)} &RandomBrightnessContrast &0.2 \\
        \textbf{(VI)} &Solarize &0.2 \\
        \textbf{(VII)} &ColorJitter &0.2 \\\bottomrule

\end{tabular}}%
    \caption{\textbf{Augmentation type}}
    \label{tab:augmentation}
\end{table}

\section*{Appendix B. Failure Cases}
\begin{figure}[t]
\centering
\includegraphics[width=0.48\textwidth]{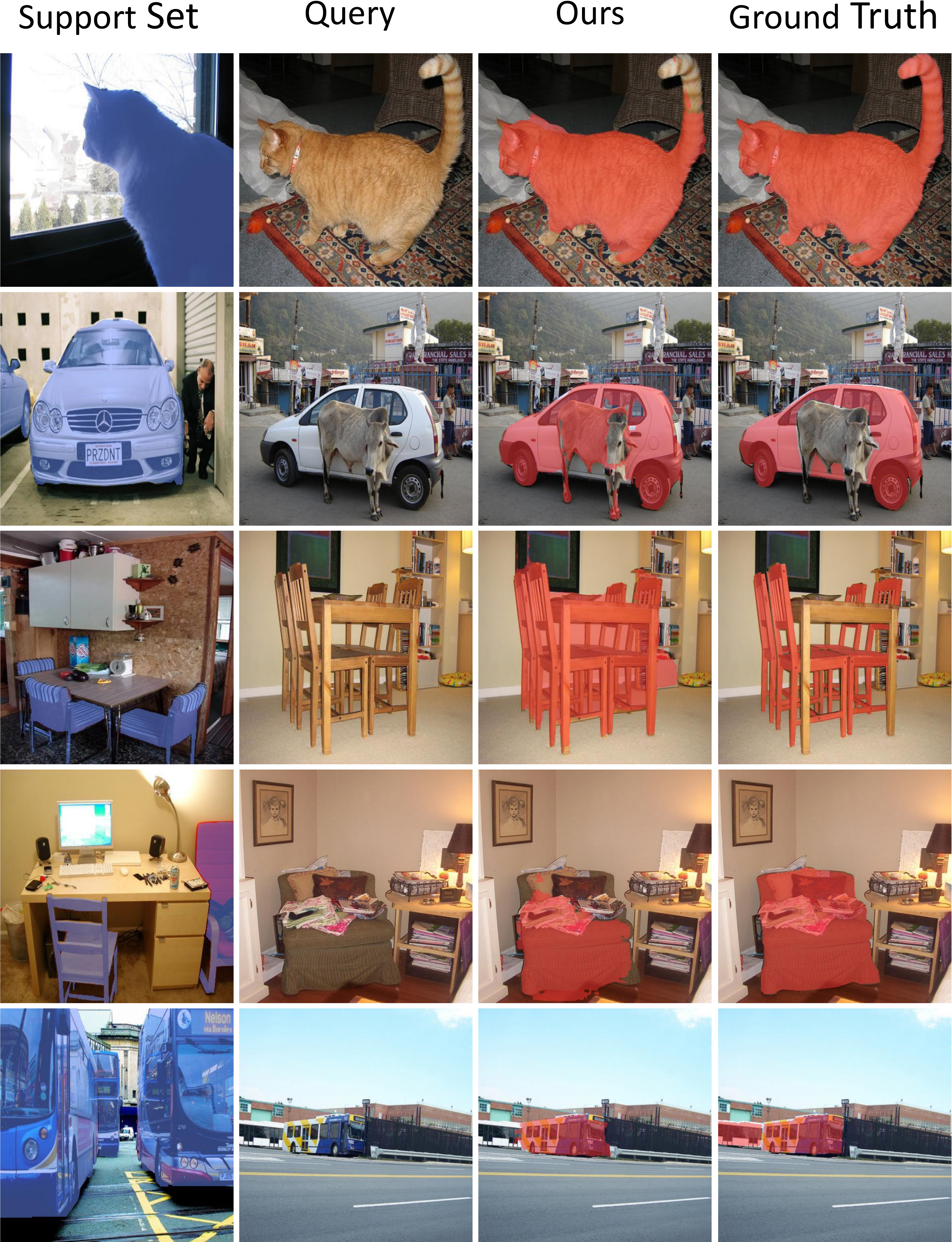}
\caption{\textbf{Failure cases.}   }
\label{Failure}
\end{figure}

Overall, we observe that our method may lack an ability to preserve fine-details and address multi-correspondence as shown in Figure~\ref{Failure}. Given an object in the support image and multiple corresponding objects in query image, VAT sometimes seem to fail finding correspondence. From these, perhaps, it could be argued that the proposed method is better at encoding matching information and effectively aggregating them than others, but it might lack an ability to better find multi-objects or preserve the fine-details. Better means to consider both factors either by introducing some confidence module~\cite{truong2021learning} or preserve fine-details when increasing the resolutions of the predicted mask during the process in decoder can be a promising direction future work. 

\section*{Appendix C. Memory and Run-time}
We additionally provide memory and run-time comparison to other aggregators and HSNet~\cite{min2021hypercorrelation}. The results are obtained using a single NVIDIA GeForce RTX 3090 GPU and Intel Core i7-10700 CPU. As shown in Table~\ref{tab:memory}, we observe that VAT is relatively slower and comsumes more memory than other aggregators. About 0.3 GB of more memory consumption and 5 ms slower run-time occur  with the proposed VAT in return for the performance. From the perspective of aggregators, 0.3 GB and 5 ms gap seem quite trivial to us.

We also provide direct comparison to the current state-of-the-art method, HSNet~\cite{min2021hypercorrelation}, in Table~\ref{tab:overallmemory}. We observe that memory consumption gap is quite large, which is an apparent limitation to the proposed method. Run-time gap is quite trivial, enabling real-time inference. Note that standard transformer can not be used due to OOM. In light of this, although  we managed to propose a method to aggregate a raw correlation map without arbitrarily changing its resolutions prior to feeding into networks, we observe relatively large memory consumption gap to HSNet~\cite{min2021hypercorrelation}, which suggests a promising direction for future work, {\it i.e.,} even more efficient method while enabling to process at higher spatial resolutions.

\section*{Appendix D. More Results}
For few-shot segmentation, VAT clearly sets new state-of-the-art for all the benchmarks, demonstrating its effectiveness. Also as VAT outperforms other SOTA methods by large margin for semantic correspondence, we argue that the cost aggregation is indeed a prime importance, even for few-shot segmentation as well. We argue that this performance boost in few-shot segmentation can be attributed to VAT's ability to find accurate semantic correspondences when tackling the segmentation task. In light of this, we are suggesting a paradigm shift for the few-shot segmentation task.
\vspace{-10pt}
\paragraph{Quantitative Results on Semantic Correspondence.}
As shown in Table~\ref{tab:spair} and Table~\ref{tab:pascal}, we provide more quantitative results on  SPair-71k~\cite{min2019spair}, PF-PASCAL~\cite{ham2017proposal}, and PF-WILLOW~\cite{ham2016proposal} in comparison to other semantic correspondence methods, including CNNGeo~\cite{rocco2017convolutional}, WeakAlign~\cite{rocco2018end}, NC-Net~\cite{Rocco18b}, HPF~\cite{min2019hyperpixel}, SFNet~\cite{lee2019sfnet}, DCC-Net~\cite{huang2019dynamic}, GSF~\cite{jeon2020guided}, SCOT~\cite{liu2020semantic}, DHPF~\cite{min2020learning}, CHM~\cite{min2021convolutional}, MMNet~\cite{zhao2021multi}, PMNC~\cite{Lee_2021_CVPR} and CATs~\cite{cho2021semantic}. \vspace{-10pt}

\begin{figure*}[t]
\centering
\includegraphics[width=0.99\textwidth]{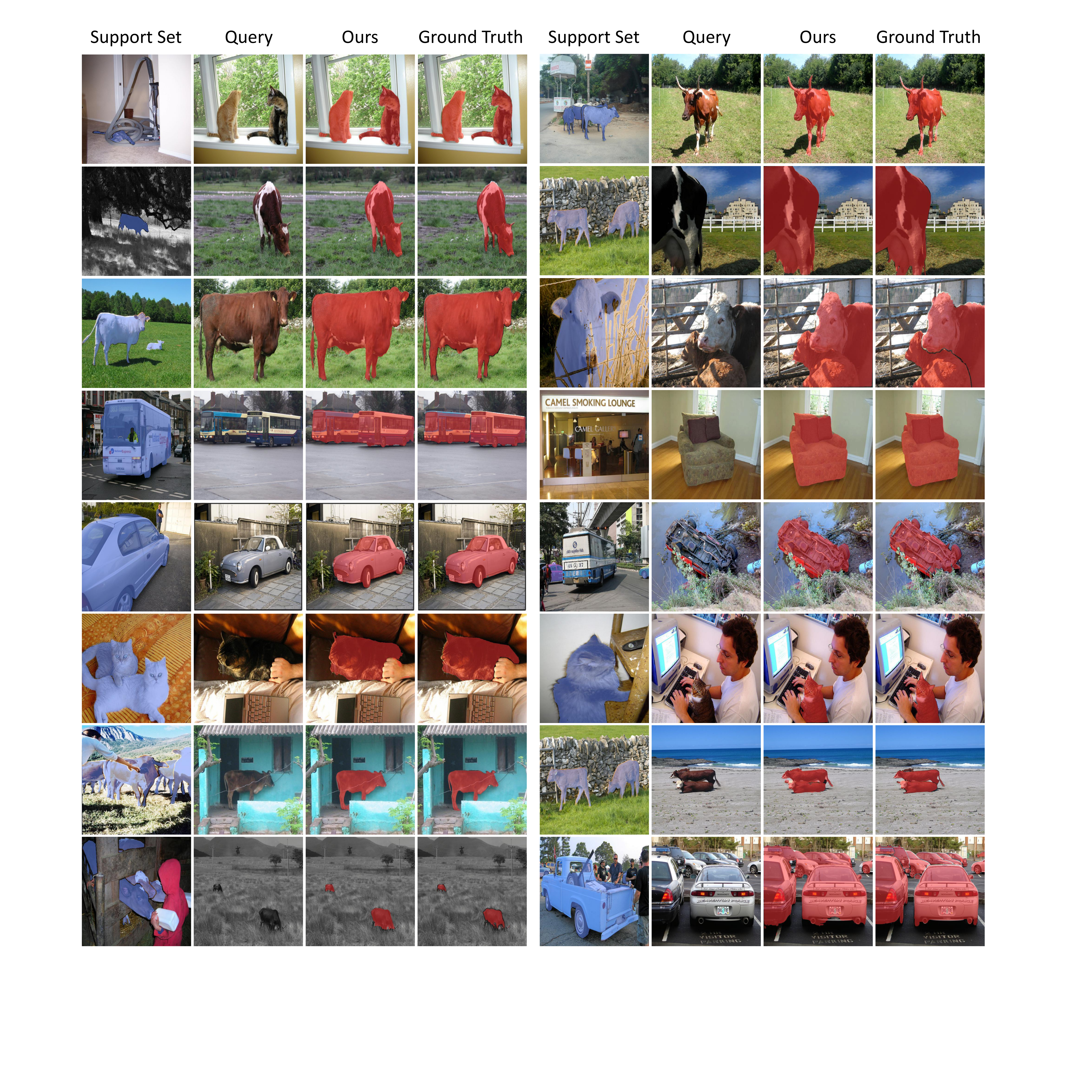}
\caption{\textbf{Qualitative results on PASCAL-5$^{i}$~\cite{shaban2017one}.}   }
\label{PASCAL}
\end{figure*}

\begin{figure*}[t]
\centering
\includegraphics[width=0.99\textwidth]{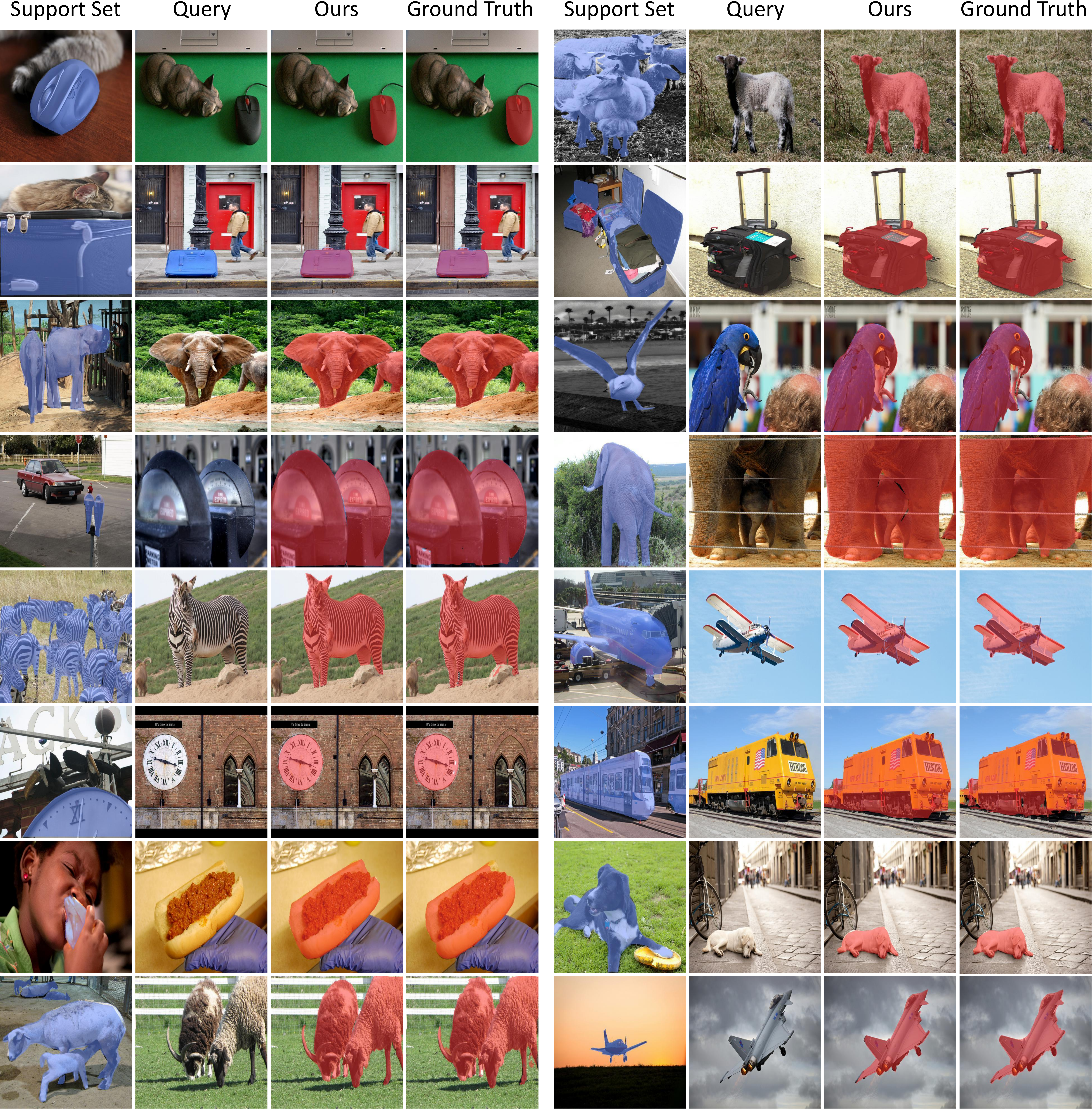}
\caption{\textbf{Qualitative results on COCO-20$^{i}$~\cite{lin2014microsoft}.}   }
\label{COCO}
\end{figure*}

\begin{figure*}[t]
\centering
\includegraphics[width=0.99\textwidth]{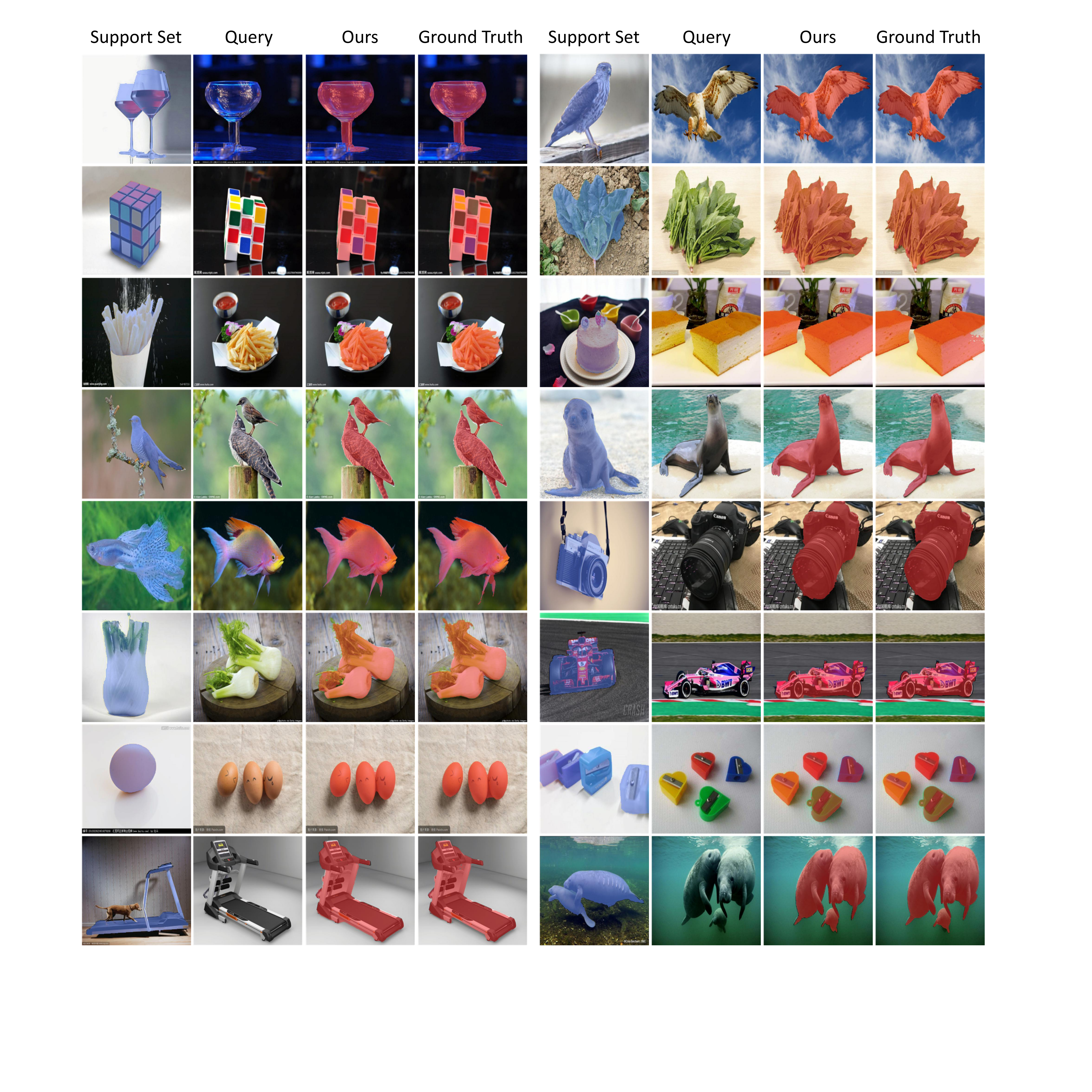}
\caption{\textbf{Qualitative results on FSS-1000~\cite{li2020fss}.}   }
\label{FSS}
\end{figure*}

\begin{figure*}[t]
\centering
\includegraphics[width=0.99\textwidth]{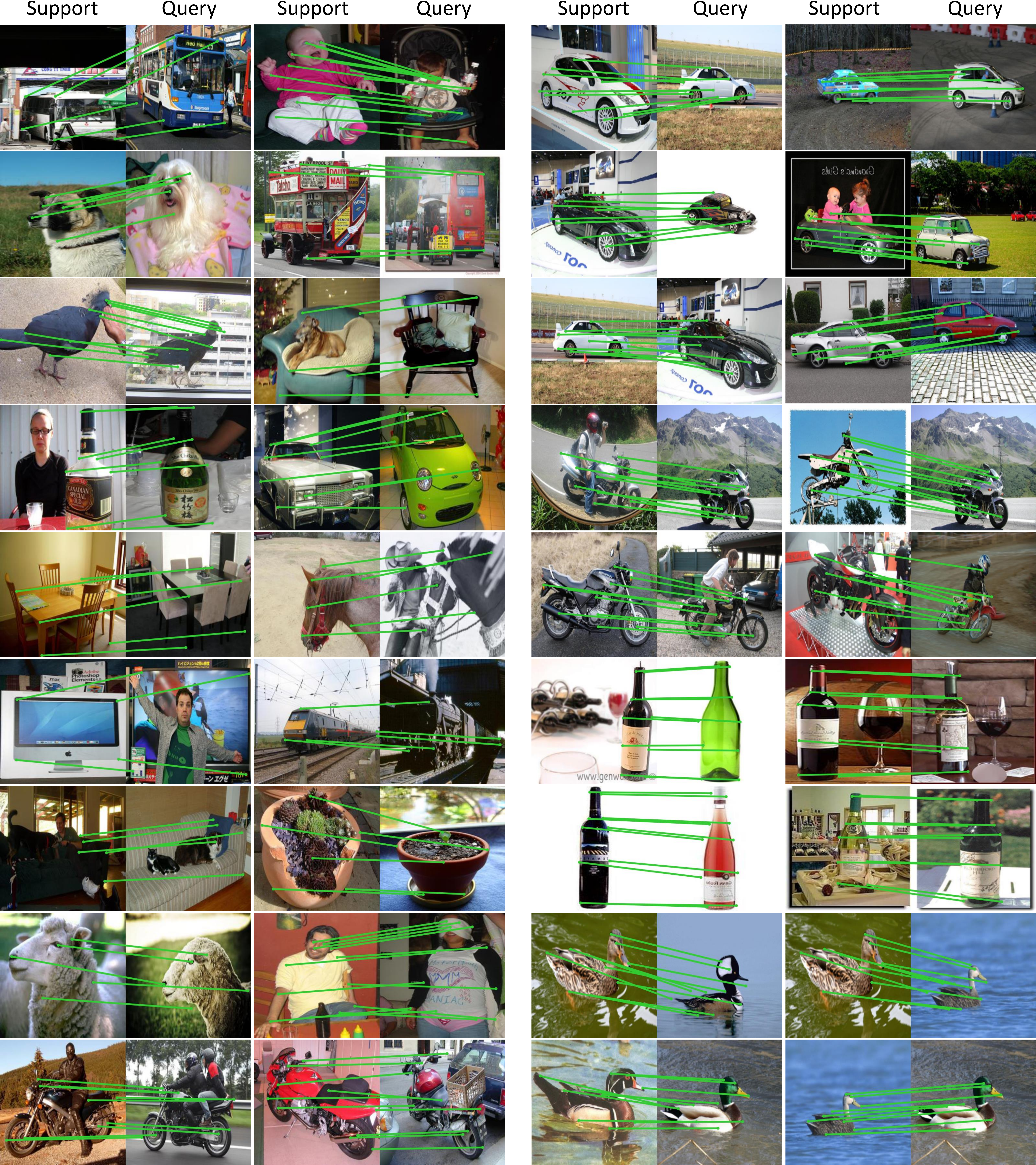}
\caption{\textbf{Qualitative results on PF-PASCAL~\cite{ham2017proposal} (left) and PF-WILLOW~\cite{ham2016proposal} (right). }   }
\label{WILLOW}
\end{figure*}

\begin{figure*}[t]
\centering
\includegraphics[width=0.8\textwidth]{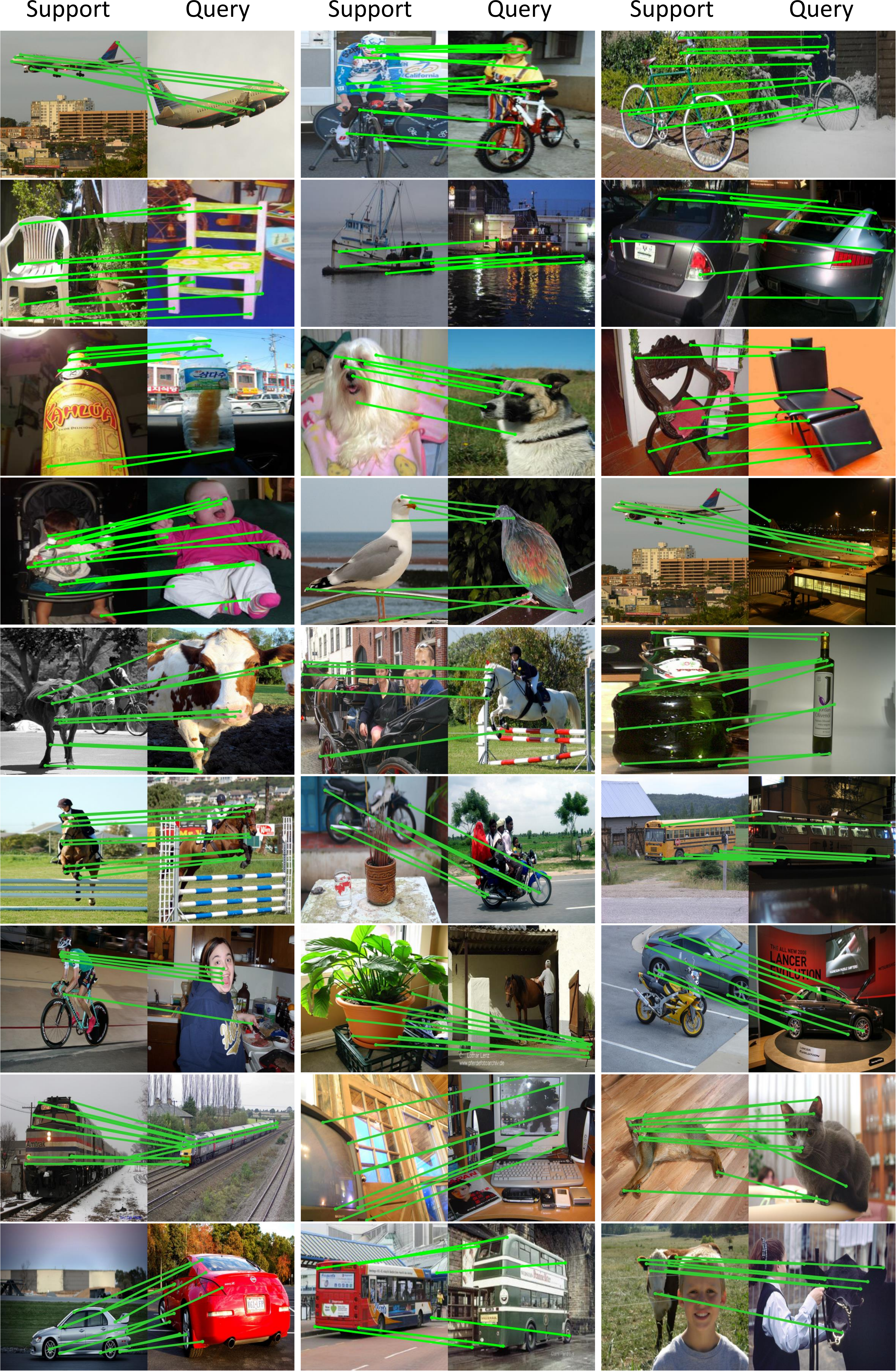}
\caption{\textbf{Qualitative results on SPair-71k~\cite{min2019spair}.}   }
\label{SPAIR}
\end{figure*}

\paragraph{Qualitative Results.}
As shown in Figure~\ref{PASCAL}, Figure~\ref{COCO}, Figure~\ref{FSS}, Figure~\ref{WILLOW} and Figure~\ref{SPAIR}, we provide qualitative results on all the benchmarks, which includes PASCAL-5$^{i}$~\cite{shaban2017one}, COCO-20$^{i}$~\cite{lin2014microsoft}, FSS-1000~\cite{li2020fss}, PF-PASCAL~\cite{ham2017proposal}, PF-WILLOW~\cite{ham2016proposal} and SPair-71k~\cite{min2019spair}.
\newpage
\newpage
\newpage
\newpage
\newpage
\clearpage
\newpage

\end{document}